\documentclass[10pt,twocolumn,letterpaper]{article}
\pdfoutput=1

\usepackage{iccv}
\makeatletter
\@namedef{ver@everyshi.sty}{}
\makeatother
\usepackage[accsupp]{axessibility}
\usepackage{pgf-pie}
\usepackage{times}
\usepackage{epsfig}
\usepackage{graphicx}
\usepackage{amsmath,amssymb}

\usepackage{booktabs} %
\usepackage{glossaries}
\usepackage[font=small,labelfont=bf]{caption}
\usepackage{subcaption}
\usepackage{multirow}
\usepackage{xcolor}
\usepackage{pifont}%
\usepackage{xspace}
\usepackage{ifthen}
\usepackage{blindtext}
\usepackage{enumitem}

\usepackage[pagebackref=true,breaklinks=true,letterpaper=true,colorlinks,bookmarks=false]{hyperref}

\newcommand{\paperversion}{NOTDRAFT} %
\iccvfinalcopy %
\def\iccvPaperID{6653} %

\ificcvfinal\fi

\newcommand{\Alg}{MDETR\xspace}
\definecolor{myblue}{rgb}{0.00392156862745098, 0.45098039215686275, 0.6980392156862745}
\definecolor{myorange}{rgb}{0.8705882352941177, 0.5607843137254902, 0.0196078431372549}
\definecolor{mygreen}{rgb}{0.00784313725490196, 0.6196078431372549, 0.45098039215686275}

\newcommand{\aish}[1]{\ifthenelse{\equal{\paperversion}{draft}}{{\color{green!40!blue}#1}}}
\newcommand{\nico}[1]{\ifthenelse{\equal{\paperversion}{draft}}{{\color{blue!20!red}[#1]}}}
\newcommand{\gab}[1]{\ifthenelse{\equal{\paperversion}{draft}}{{\color{blue}[\textbf{Gab}:#1]}}}
\newcommand{\mannat}[1]{\ifthenelse{\equal{\paperversion}{draft}}{{\color{orange}[\textbf{Mannat}:#1]}}}
\newcommand{\ishan}[1]{\ifthenelse{\equal{\paperversion}{draft}}{{\color{magenta}[\textbf{Ishan}:#1]}}}

\author{Aishwarya Kamath$^1$ ~ Mannat Singh$^2$ ~
Yann LeCun$^{123}$ ~ Gabriel Synnaeve$^2$ ~ Ishan Misra$^2$ \\ Nicolas Carion$^3$ \\
$^1$NYU Center for Data Science ~ $^2$Facebook AI Research ~ $^3$NYU Courant Institute \\
{\tt\small \{aish, yann.lecun, nc2794\}@nyu.edu, \{mannatsingh,imisra,gab\}@fb.com}
}

\title{MDETR - Modulated Detection for End-to-End Multi-Modal Understanding}

\begin{document}

\maketitle

\begin{abstract}
Multi-modal reasoning systems rely on a pre-trained object detector to extract regions of interest from the image. However, this crucial module is typically used as a black box, trained independently of the downstream task and on a fixed vocabulary of objects and attributes.
This makes it challenging for such systems
to capture the long tail of visual concepts expressed in free form text.
In this paper we propose MDETR, an end-to-end \emph{modulated} detector that detects objects in an image conditioned on a raw text query, like a caption or a question.
We use a transformer-based architecture to reason jointly over text and image by fusing the two modalities at an early stage of the model.
We pre-train the network on 1.3M text-image pairs, mined from pre-existing multi-modal datasets having explicit alignment between phrases in text and objects in the image.
We then fine-tune on several downstream tasks such as phrase grounding, referring expression comprehension and segmentation, achieving state-of-the-art results on popular benchmarks.
We also investigate the utility of our model as an object detector on a given label set when fine-tuned in a few-shot setting. We show that our pre-training approach provides a way to handle the long tail of object categories which have very few labelled instances. Our approach can be easily extended for visual question answering, achieving competitive performance on GQA and CLEVR. The code and models are available at \url{https://github.com/ashkamath/mdetr}.

\end{abstract}
\setlength{\tabcolsep}{5pt}

\section{Introduction}
\label{sec:introduction}
Object detection forms an integral component of most state-of-the-art multi-modal understanding systems \cite{chen2019uniter, li2020oscar}, typically used as a black-box to detect a fixed vocabulary of concepts in an image followed by multi-modal alignment. This ``pipelined'' approach limits co-training with other modalities as context and restricts the downstream model to only have access to the detected objects and not the whole image. In addition, the detection system is usually frozen, which prevents further refinement of the model's perceptive capability. In the vision-language setting, it implies restricting the vocabulary of the resulting system to the categories and attributes of the detector, and is often a  bottleneck for performance on these tasks~\cite{zhang2021vinvl}. As a result, such a system cannot recognize novel combinations of concepts expressed in free-form text.

A recent line of work~\cite{yang2019fast,plummer2020revisiting, hinami2017discriminative} considers the problem of text-conditioned object detection. These methods extend mainstream one-stage and two-stage detection architectures to achieve this goal.
However, to the best of our knowledge, it has not been demonstrated that such detectors can improve performance on downstream tasks that require reasoning over the detected objects, such as visual question answering (VQA).
We believe this is because these detectors are not end-to-end differentiable and thus cannot be trained in synergy with downstream tasks.

\begin{figure}[t]
        \centering
        \includegraphics[trim={0 0 0 1.42in}, clip, height=1.42in]{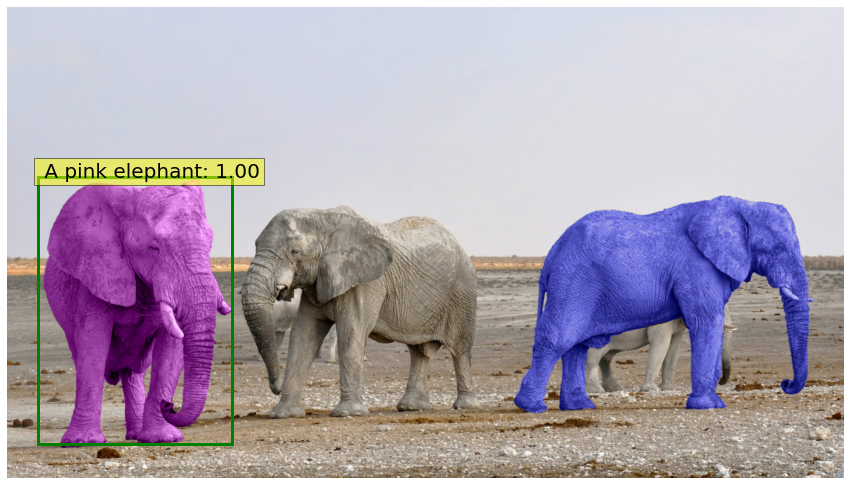}
        \caption{Output of MDETR for the query ``A pink elephant''. The colors are not segmentation masks but the real colors of the pixels. The model has never seen a pink nor a blue elephant in training.\label{fig:intropink}}
        \vspace{-0.5cm}
    \end{figure}

\begin{figure*}[t]
 \centering
  \includegraphics[width=0.96\textwidth]{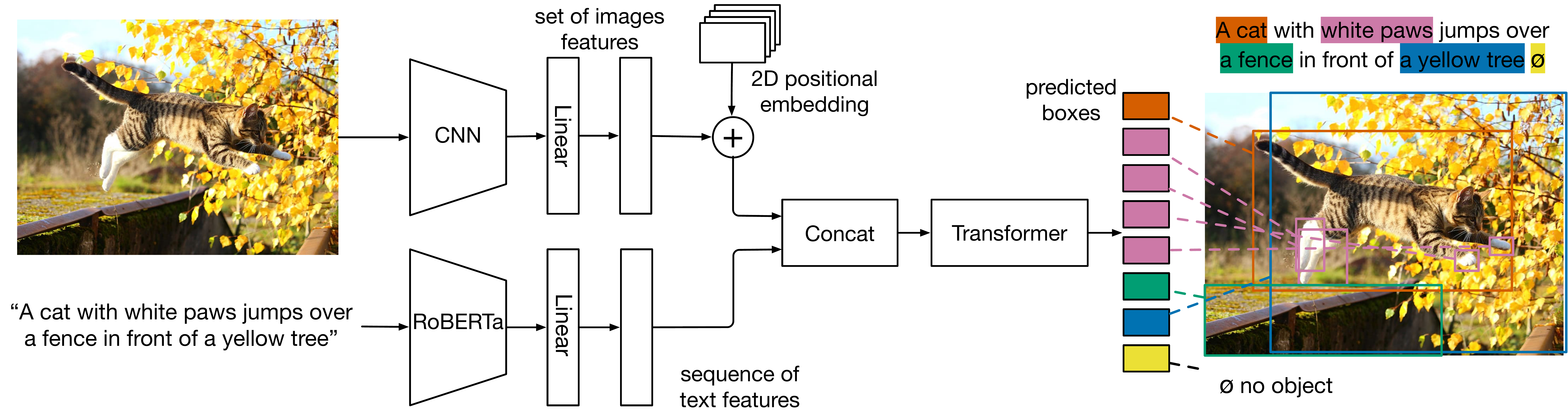}
 \caption{MDETR uses a convolutional backbone to extract visual features, and a language model such as RoBERTa to extract text features. The features of both modalities are projected to a shared embedding space, concatenated and fed to a transformer encoder-decoder that predicts the bounding boxes of the objects and their grounding in text.}
 \label{fig:model diagram}
\end{figure*}
    
Our method, \Alg, is an end-to-end \emph{modulated} detector based on the recent DETR \cite{carion2020end} detection framework, and performs object detection in conjunction with natural language understanding, enabling truly end-to-end multi-modal reasoning.
\Alg relies solely on text and aligned boxes as a form of supervision for concepts in an image. Thus, unlike current detection methods, \Alg detects nuanced concepts from free-form text, and generalizes to unseen combinations of categories and attributes. We showcase such a combination as well as modulated detection in Fig.~\ref{fig:intropink}.
By design, our predictions are grounded in text, which is a key requirement for visual reasoning~\cite{yangObjectCentricDiagnosisVisual2020}.
When pre-trained using a dataset of 200,000 images and aligned text with box annotations, we achieve best reported results on the Flickr30k  dataset for phrase grounding, RefCOCO/+/g datasets for referring expression comprehension, and referring expression segmentation on PhraseCut, as well as competitive performance on the GQA and CLEVR benchmarks for visual question answering.

Our contributions are as follows:
\begin{itemize}
    \item We introduce an end-to-end text-modulated detection system derived from the DETR detector.
    \item We demonstrate that the modulated detection approach can be applied seamlessly to solve tasks such as phrase grounding and referring expression comprehension, setting new state of the art performance on both these tasks using datasets having synthetic as well as real images.
    \item We show that good modulated detection performance naturally translates to downstream task performance, for instance achieving competitive performance on visual question answering, referring expression segmentation, and on few-shot long-tailed object detection. 
    
\end{itemize}

\section{Method}
\label{sec:method}
In this section we first briefly summarize the object detection pipeline \cite{carion2020end} based on which we build our model in \S \ref{subsec:prelim} and then describe how we extend it for modulated detection in \S \ref{subsec:mdetr}.

\subsection{Background}
\label{subsec:prelim}
\paragraph{DETR} \label{subsubsec:detr}
Our approach to modulated detection builds on the DETR system \cite{carion2020end}, which we briefly review here. We refer the readers to the original paper for additional details.
DETR is an end-to-end detection model composed of a backbone (typically a convolutional residual network \cite{he2016deep}), followed by a Transformer Encoder-Decoder \cite{Vaswani2017AttentionIA}.

The DETR encoder operates on 2D flattened image features from the backbone and applies a series of transformer layers. The decoder takes as input a set of $N$ learned embeddings called \emph{object queries}, that can be viewed as slots that the model needs to fill with detected objects. All the object queries are fed in parallel to the decoder, which uses cross-attention layers to look at the encoded image and predicts the output embeddings for each of the queries. The final representation of each object query is independently decoded into box coordinates and class labels using a shared feed-forward layer.
The number of object queries acts as a de facto upper-bound on the number of objects the model can detect simultaneously. It has to be set to a sufficiently large upper-bound on the number of objects one may expect to encounter in a given image. Since the actual number of objects in a particular image may be less than the number of queries $N$, an extra class label corresponding to ``no object" is used, denoted by $\varnothing$. The model is trained to output this class for every query that doesn't correspond to an object. 

DETR is trained using a Hungarian matching loss, where a bipartite matching is computed between the $N$ proposed objects and the ground-truth objects. Each matched object is supervised using the corresponding target as ground-truth, while the un-matched objects are supervised to predict the ``no object" label $\varnothing$. The classification head is supervised using standard cross-entropy, while the bounding box head is supervised using a combination of absolute error (L1 loss) and Generalized IoU \cite{rezatofighi2019generalized}.

\subsection{MDETR}
\label{subsec:mdetr}
\subsubsection{Architecture}
We depict the architecture for MDETR in Fig.~\ref{fig:model diagram}.
As in DETR, the image is encoded by a convolutional backbone and flattened. In order to conserve the spatial information, 2-D positional embeddings are added to this flattened vector. We encode the text using a pre-trained transformer language model to produce a sequence of hidden vectors of same size as the input. We then apply a modality dependent linear projection to both the image and text features to project them into a shared embedding space. These feature vectors are then concatenated on the sequence dimension to yield a single sequence of image and text features.
This sequence is fed to a joint transformer encoder termed as the \textit{cross encoder}. Following DETR, we apply a transformer decoder on the object queries while cross attending to the final hidden state of the cross encoder.
The decoder's output is used for predicting the actual boxes.
\vspace{-0.25cm}

\subsubsection{Training}
We present the two additional loss functions used by \Alg, which encourage alignment between the image and the text. Both of these use the same source of annotations: free form text with aligned bounding boxes. The first loss function that we term as the \textit{soft token prediction} loss is a non parametric alignment loss. The second, termed as the \textit{text-query contrastive alignment} is a parametric loss function enforcing similarity between aligned object queries and tokens. 

\vspace{0.05in}
\par \noindent \textbf{Soft token prediction} 
\label{soft-token-prediction}
For modulated detection, unlike in the standard detection setting, we are not interested in predicting a categorical class for each detected object. 
Instead, we predict the span of tokens from the original text that refers to each matched object. Concretely, we first set the maximum number of tokens for any given sentence to be L~$=$~256. For each predicted box that is matched to a ground truth box using the bi-partite matching, the model is trained to predict a uniform distribution over all \textit{token positions} that correspond to the object.
Fig.~\ref{fig:model diagram} shows an example where the box for cat is trained to predict a uniform distribution over the first two words.
In Fig.~\ref{fig:soft_loss}, we show a simplified visualization of the loss for this example, in terms of a distribution over words for each box, but in practice we use token spans after tokenization using a BPE scheme \cite{sennrich2015neural}. 
Any query that is not matched to a target is trained to predict the ``no object" label $\varnothing$. 
Note that several words in the text could correspond to the same object in the image, and conversely several objects could correspond to the same text. For example, ``a couple'' referred to by two boxes in the image, could further be referred to individually in the same caption. By designing the loss function in this way, our model is able to learn about co-referenced objects from the same referring expression.

\textbf{Contrastive alignment} 
While the soft token prediction uses \textit{positional} information to align the objects to text, the contrastive alignment loss enforces alignment between the \textit{embedded representations} of the object at the output of the decoder, and the text representation at the output of the cross encoder. 
This additional contrastive alignment loss ensures that the embeddings of a (visual) object and its corresponding (text) token are closer in the feature space compared to embeddings of unrelated tokens.
This constraint is stronger than the soft token prediction loss as it directly operates on the representations and is not solely based on positional information.
More concretely, consider the maximum number of tokens to be $L$ and maximum number of objects to be $N$. Let $T^+_i$ be the set of tokens that a given object $o_i$ should be aligned to, and $O^+_i$ be the set of objects to be aligned with a given token $t_i$. \\
The contrastive loss for all objects, inspired by InfoNCE \cite{Oord2018RepresentationLW} is normalized by number of positive tokens for each object and can be written as follows:
\begin{equation}
l_{o} = \sum_{i=0}^{N-1} \frac{1}{\vert T^+_i \vert } \sum_{j \in T^+_i}  - \log \bigg( \frac{\exp(o_i^\top t_j / \tau)}{\sum_{k=0}^{L-1} \exp(o_i^\top t_k/\tau) } \bigg)    
\end{equation}
where $\tau$ is a temperature parameter that we set to 0.07 following literature \cite{wu2018unsupervised, radford2021learning}. By symmetry, the contrastive loss for all tokens, normalized by the number of positive objects for each token is given by:
\begin{equation}
l_{t} = \sum_{i=0}^{L-1} \frac{1}{\vert O^+_i \vert} \sum_{j \in O^+_i} - \log \bigg( \frac{\exp( t_i^\top o_j / \tau)}{\sum_{k=0}^{N-1} \exp(t_i^\top o_k /\tau) } \bigg)    
\end{equation}
We take a the average of these two loss functions as our contrastive alignment loss. 

\textbf{Combining all the losses}
In MDETR, a bipartite matching is used to find the best match between the predictions and the ground truth targets just as in DETR. The main difference is that there is no class label predicted for each object - instead predicting a uniform distribution over the relevant positions in the text that correspond to this object (soft token predictions), supervised using a soft cross entropy.
The matching cost consists of this in addition to the L1 \& GIoU loss between the prediction and the target box as in DETR. After matching, the total loss consists of the box prediction losses (L1 \& GIoU), soft-token prediction loss, and the contrastive alignment loss.

\label{subsubsec:training_obj}

\section{Experiments}
\label{sec:experiments}
In this section we describe the data and training used for pre-training MDETR, and provide details and results on the tasks that we use to evaluate our approach. Results on the CLEVR dataset are reported in Table \ref{tab:clevrresults_small}. For a discussion on the CLEVR results and further details on data preparation and training, please see Appendix \ref{appendix_section:CLEVR_results_detailed}. Experimental details for pre-training and downstream tasks on natural images are detailed in \S \ref{subsec:pretraining} and  \S \ref{subsec:downstream}.

\begin{table}[t]
\setlength{\tabcolsep}{1.75pt}
\begin{center}
\small
 \begin{tabular}{ccccccccc} 
 \toprule
Method & {CLEVR} & \multicolumn{2}{c}{CLEVR-Hu} & \multicolumn{2}{c}{CoGenT} &CLEVR-Ref+ \\ [0.5ex]
        & Overall & - FT & + FT & TestA & TestB& Acc  \\
 \midrule 
 MAttNet\cite{yu_mattnet_2018}&-&-&-&-&-&60.9\\
 MGA-Net\cite{Zheng_2020_ACCV}&-&-&-&-&-&80.1\\
 FiLM\cite{perezFiLMVisualReasoning2017} &97.7 &56.6 & 75.9 & 98.3 & \textbf{78.8}&-\\

 MAC \cite{Hudson2018CompositionalAN}   & 98.9  & 57.4 & 81.5 & - & - &-\\
 NS-VQA\cite{yiNeuralSymbolicVQADisentangling2019}$^*$ & \textbf{99.8} & -&67.8&\textbf{99.8} & 63.9 &-\\
 OCCAM \cite{wangInterpretableVisualReasoning2020} & 99.4 & - & - & - & - &- \\
 MDETR & 99.7  & \textbf{59.9} & \textbf{81.7} & \textbf{99.8} & 76.7 &\textbf{100}\\
\bottomrule
\end{tabular}
\caption{Results on CLEVR-based datasets. We report accuracies on the test set of CLEVR. On CLEVR-Humans, we report accuracy on the test set before and after fine-tuning. On CoGenT, we report performance when the model is trained in condition A, without finetuning on condition B. On CLEVR-Ref+, we report the accuracy on the subset where the referred object is unique. *indicates method uses external program annotations. Further details in Appendix \ref{appendix_section:CLEVR_results_detailed}.} 
\label{tab:clevrresults_small} \vspace{-0.75cm}
\end{center}
\end{table}

\subsection{Pre-training Modulated Detection}
\label{subsec:pretraining}
For pre-training, we focus on the task of modulated detection where the aim is to detect all the objects that are referred to in the aligned free form text. We create a combined dataset using images from the Flickr30k \cite{plummer2015flickr30k}, MS COCO \cite{Lin2014MicrosoftCC} and Visual Genome (VG)  \cite{krishnavisualgenome} datasets. Annotations from the referring expressions datasets, VG regions, Flickr entities and GQA train balanced set are used for training. An image may have several text annotations associated with it. Details on the datasets can be found in Appendix~\ref{sec:appendix_datasets}.

\begin{table*}[t]
\setlength{\tabcolsep}{5pt}
\begin{center}
\small
 \begin{tabular}{ccccccccccc} 
 \toprule
 Method & Detection & Pre-training & \multicolumn{3}{c}{RefCOCO} &  \multicolumn{3}{c}{RefCOCO+} & \multicolumn{2}{c}{RefCOCOg}  \\ [0.5ex] 
       & backbone  & image data  & val & testA & testB & val & testA & testB & val & test  \\
 \midrule

 MAttNet\cite{yu_mattnet_2018} & R101 & None  &76.65  & 81.14 & 69.99 & 65.33 & 71.62 & 56.02 & 66.58 & 67.27  \\
 ViLBERT\cite{lu2019vilbert}  & R101 & CC (3.3M) & - & -  & - & 72.34 & 78.52 &  62.61 & - & -   \\ 
 VL-BERT\_L \cite{su2019vl} & R101 & CC (3.3M) &- & -  & - & 72.59 & 78.57 &  62.30 & - & -\\
 UNITER\_L\cite{chen2019uniter}$^*$ & R101 & CC, SBU, COCO, VG (4.6M) & 81.41 & 87.04  & 74.17 & 75.90  & 81.45 & 66.70 & 74.86 & 75.77 \\  
 VILLA\_L\cite{gan2020large}$^*$ & R101 & CC, SBU, COCO, VG (4.6M) & 82.39 & 87.48  & 74.84 &  76.17 & 81.54 &  66.84 & 76.18 & 76.71   \\ 
 ERNIE-ViL\_L\cite{yu2020ernie} & R101 & CC, SBU (4.3M) &- & -  & - & 75.95 & 82.07 &  66.88 & - & -  \\  
 MDETR & R101 & COCO, VG, Flickr30k (200k)  & \textbf{86.75} & \textbf{89.58}& \textbf{81.41}    & \textbf{79.52}  & \textbf{84.09} &  \textbf{70.62}  & \textbf{ 81.64} &\textbf{ 80.89}  \\
 MDETR & ENB3 & COCO, VG, Flickr30k (200k)  & \textbf{87.51} & \textbf{90.40 }& \textbf{82.67}    & \textbf{81.13}  & \textbf{85.52} &  \textbf{ 72.96}  & \textbf{ 83.35} &\textbf{ 83.31}  \\
 \bottomrule
\end{tabular}
\caption{Accuracy results on referring expression comprehension. *As mentioned in UNITER \cite{chen2019uniter}, methods using box proposals from the BUTD detector \cite{anderson2018bottom} suffer from a test set leak, since the detector was trained on images including the validation and test set of the RE comprehension datasets. We report numbers for these methods from their papers using these ``contaminated features" but we would like to stress that all of our pre-training excluded the images used in the val/test of any of the downstream datasets including for RE comprehension. CC refers to Conceptual Captions \cite{sharma-etal-2018-conceptual}, VG to Visual Genome \cite{krishnavisualgenome}, SBU refers to the SBU Captions\cite{ordonez2011im2text} and COCO to Micosoft COCO \cite{Lin2014MicrosoftCC}.}
\label{tab:refexp}
\end{center}
\end{table*}

\begin{figure}
 \centering
 \includegraphics[height=3.0in]{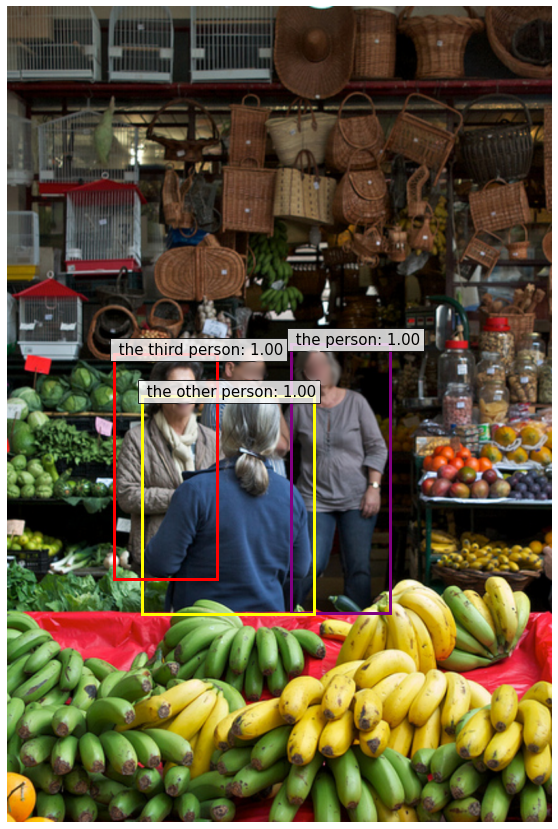}
 \caption{Our combination of annotations results in examples such as the following: ``the person in the grey shirt with a watch on their wrist. the other person wearing a blue sweater. the third person in a gray coat and scarf." We show the predictions from our model for this caption. It is able to pay attention to all the objects in the image and then disambiguate between them based on the text. The model is trained to predict the root of the phrase as the positive token span, which as we can see in this figure, correctly refers to the three different people.}
 \label{fig:refexp}\vspace{-0.5cm}
\end{figure}

\textbf{Data combination} For each image, we take all annotations from these datasets and combine the text that refers to the same image while ensuring that all images that are in the validation or testing set for all our downstream tasks are removed from our train set. The combination of sentences is done using a graph coloring algorithm which ensures that only phrases having boxes with GIoU $\leq$ 0.5 are combined, and that the total length of a combined sentence is less than 250 characters. In this way, we arrive at a dataset having 1.3M aligned image - text pairs. This combination step is important for two reasons: 1) data efficiency, by packing more information into a single training example and 2) it provides a better learning signal for our soft token prediction loss since the model has to learn to disambiguate between multiple occurrences of the same object category, as depicted in Fig \ref{fig:refexp}. In the single sentence case, the soft token prediction task becomes trivial since it can always predict the root of the sentence without looking at the image. Experimentally, we find that such dense annotations translate to better grounding between text and image and subsequently to better downstream performance. 

\setlength{\tabcolsep}{3.5pt}
\begin{table}[t]
\begin{center}
\small
\begin{tabular}{ccccccc}
 \toprule
 Method & \multicolumn{3}{c}{Val} &  \multicolumn{3}{c}{Test} \\
  & R@1 & R@5 & R@10  & R@1 & R@5 & R@10 \\
 \midrule
 &  \multicolumn{6}{c}{\textsc{Any-Box-Protocol}}  \\
 \midrule
 BAN \cite{kim2018bilinear} & - &  - & - & 69.7 &  84.2 &   86.4  \\
  VisualBert\cite{li2019visualbert} &  68.1 &  84.0 & 86.2 & - & - &  - \\
  VisualBert$\dagger$\cite{li2019visualbert}&  70.4 &  84.5 & 86.3 & 71.3 &  85.0 &  86.5  \\
  MDETR-R101 & 78.9  & 88.8 & 90.8 & - & - & - \\ 
  MDETR-R101$\dagger*$ & \textbf{82.5} & \textbf{92.9} & \textbf{94.9}  & \textbf{83.4} & \textbf{93.5} & \textbf{95.3} \\
  MDETR-ENB3$\dagger*$  &  \textbf{82.9} & \textbf{93.2} & \textbf{95.2}  & \textbf{84.0} & \textbf{93.8} & \textbf{95.6} \\
  MDETR-ENB5$\dagger*$  &  \textbf{83.6} & \textbf{93.4} & \textbf{95.1}  & \textbf{84.3} & \textbf{93.9} & \textbf{95.8} \\
   \midrule
   & \multicolumn{6}{c}{\textsc{Merged-Boxes-Protocol}}  \\
   \midrule
   CITE \cite{plummer2018conditional} & - &  - & - & 61.9 &  - &   -  \\
   FAOG \cite{yang2019fast} & - &  - & - & 68.7 &  - &   -  \\
   SimNet-CCA \cite{plummer2020revisiting}  & - &  - & - & 71.9 &  - &   -  \\
   DDPN \cite{yu2018rethining}  & 72.8 &  - & - & 73.5 &  - &   -  \\
    MDETR-R101 & 79.0 & 86.7 & 88.6  & - & - & - \\
    MDETR-R101$\dagger*$ & \textbf{82.3} & \textbf{91.8} & \textbf{93.7}  & \textbf{83.8} & \textbf{92.7} & \textbf{94.4} \\
 \bottomrule
\end{tabular}
\caption{Results on the phrase grounding task on Flickr30k entities dataset \cite{plummer2015flickr30k}. Models with $\dagger$ are pre-trained on COCO, models with $*$ are also pre-trained on VG and Flickr 30k. Our models (MDETR) use a RoBERTa text encoder while other models use RNNs, word2vec-based features, or BERT (comparable to RoBERTa) text encoders. All models use a ResNet101 backbone, except MDETR-ENB3 which uses EfficientNet-B3 and  MDETR-ENB5 with an EfficientNet-B5.\vspace{-0.5cm}
\label{tab:flickr}}
\end{center}
\end{table}
\setlength{\tabcolsep}{5pt}

\textbf{Model}
We use a pre-trained RoBERTa-base \cite{liu2019roberta} as our text encoder, having 12 transformer encoder layers, each with hidden dimension of 768 and 12 heads in the multi-head attention. We use the implementation and weights from HuggingFace \cite{Wolf2020TransformersSN}.
For the visual backbone, we explore two options. 
The first is a ResNet-101~\cite{he2016deep} pretrained on ImageNet with frozen batchnorm layers, taken from Torchvision. 
This is to be comparable with current literature in the space of multi-modal understanding, where the popular approach is to use the BUTD object detector with a Resnet-101 backbone from \cite{anderson2018bottom} trained on the VG dataset. 
In our work, we are not limited by the existence of pre-trained detectors, and inspired by its success in object detection \cite{tanEfficientDetScalableEfficient2020}, we choose to explore the EfficientNet family~\cite{tanEfficientNetRethinkingModel2020} for our backbone. We use a model which was trained on large amounts of unlabelled data in addition to ImageNet, using a pseudo-labelling technique called Noisy-Student~\cite{xieSelftrainingNoisyStudent2020}. We choose the EfficientNetB3, which achieves 84.1\% top 1 accuracy on ImageNet with only 12M weights and EfficientB5 which achieves 86.1\% using 30M weights. We use the implementation provided by the Timm library~\cite{rw2019timm}, and freeze the batchnorm layers. We pre-train our model for 40 epochs on 32 V100 gpus with an effective batch size of 64, which takes approximately a week to train. Training hyperparameters are detailed in Appendix \ref{sec:appendix_hyperparameters}.

\subsection{Downstream Tasks}
\label{subsec:downstream}
We evaluate our method on 4 downstream tasks: referring expression comprehension and segmentation, visual question answering and phrase grounding. Training hyperprameters for all tasks can be found in Appendix \ref{sec:appendix_hyperparameters}.

\textbf{Phrase grounding}
Given one or more phrases, which may be inter-related, the task is to provide a set of bounding boxes for each phrase.  We use the Flickr30k entities dataset for this task, with the train/val/test splits as provided by \cite{plummer2015flickr30k} and evaluate our performance in terms of Recall@k. For each sentence in the test set, we predict 100 bounding boxes and use the soft token alignment prediction to rank the boxes according to the score given to the token positions that correspond to the phrase. 
We evaluate under two protocols which we name \textsc{Any-Box} \cite{li2019visualbert, kim2018bilinear} and \textsc{Merged-Boxes} \cite{ plummer_phrase_2017}. Please see Appendix \ref{sec:eval_GD} for a discussion on the two protocols.  
We compare our method to existing state-of-the-art results from two types of approaches - the text conditioned detection models \cite{ plummer2020revisiting, yang2019fast} and a transformer based vision-language pre-training model \cite{li2019visualbert}. In the \textsc{Any-Box} setting, we obtain a 8.5 point boost over current state of the art on this task as measured in terms of Recall@1 on the validation set, without using any pre-training (no additional data). With pre-training, we further obtain a 12.1 point boost over the best model's performance on the test set, while using the same backbone.

\textbf{Referring expression comprehension}
Given an image and a referring expression in plain text, the task is to localize the object being referred to by returning a bounding box around it. The approach taken by most prior work \cite{yu_mattnet_2018, lu2019vilbert, chen2019uniter, yu2020ernie} on this task has been to rank a set of pre-extracted bounding boxes associated with an image, that are obtained using a pre-trained object detector. In this paper, we solve a much harder task - we train our model to directly \textit{predict} the bounding box, given a referring expression and the associated image. There are three established datasets for this task called RefCOCO, RefCOCO+ \cite{yu_modeling_2016} and RefCOCOg \cite{mao_generation_2016}.  Since during pre-training we annotate every object referred to within the text, there is a slight shift in the way the model is used in this task. For example, during pre-training, given the caption ``The woman wearing a blue dress standing next to the rose bush.", MDETR would be trained to predict boxes for all referred objects such as the woman, the blue dress and the rose bush. However, for referring expressions, the task would be to only return \emph{one} bounding box, which signifies the woman being referred to by the entire expression. For this reason, we finetune the model on the task specific dataset for 5 epochs. At inference time, we use the $\varnothing$ label to rank the 100 detected boxes. Let $P(\varnothing)$ be the probability assigned to the ``no object" label, we rank by decreasing order of $1-P(\varnothing)$.
We report results in Table \ref{tab:refexp}, showing large improvements over state-of-the-art across all datasets. 

\setlength{\tabcolsep}{3pt}
\begin{table}[t]
\begin{center}
\small
 \begin{tabular}{cccccc} 
 \toprule
Method & Backbone  & \multicolumn{4}{c}{PhraseCut}  \\ [0.5ex]
       &        & M-IoU & Pr@0.5 & Pr@0.7 & Pr@0.9 \\
 \midrule
 RMI\cite{Chen2019RMI} & R101 & 21.1 &  22.0 & 11.6 & 1.5 \\
 HULANet\cite{wu_phrasecut_2020} & R101 & 41.3 &  42.4 & 27.0 & 5.7 \\
 MDETR  & R101 & \textbf{53.1} & \textbf{56.1} & \textbf{38.9} &\textbf{11.9}  \\
 MDETR  & ENB3 & \textbf{53.7} & \textbf{57.5} & \textbf{39.9} &\textbf{11.9}  \\
\bottomrule
\end{tabular}
\caption{Following \cite{wu_phrasecut_2020}, we report the mean intersection-over-union (IoU) of our masks with the ground-truth masks. We also report the precision Pr@$I$ of our model, where success is marked when our proposed mask has an IoU with the ground-truth higher than the threshold $I$. With a comparable ResNet backbone, we observe consistent gains across all metrics over HULANet~\cite{wu_phrasecut_2020}, the current state-of-the-art. The EfficientNet backbone further improves on those results.}
\vspace{-0.8cm}
\label{tab:phrasecut}
\end{center}
\end{table}
\setlength{\tabcolsep}{5pt}

\textbf{Referring expression segmentation}
Similarly to DETR, we show that our approach can be extended to perform segmentation by evaluating on the referring expression segmentation task of the recent PhraseCut \cite{wu_phrasecut_2020} dataset which consists of images from VG, annotated with segmentation masks for each referring expression. These expressions comprise a wide vocabulary of objects, attributes and relations, making it a challenging benchmark. Contrary to other referring expression segmentation datasets, in PhraseCut the expression may refer to several objects. The model is expected to find all the corresponding instances.
Our training occurs in two stages. In the first step, we take our pre-trained model after 40 epochs and fine-tune it for 10 epochs on this dataset, supervising the model to output correct boxes for the referred expressions. We use the box AP on the validation set for early stopping. In the second stage, following \cite{carion2020end}, we freeze the weights of the network, and only train a segmentation head for 35 epochs, with a learning rate drop at 25 epochs, supervised using a combination of the Dice/F1 loss\cite{DICE2016} and the Focal loss \cite{linFocalLossDense2017}. At inference-time, we assign a confidence to each predicted box equal to $1-P(\varnothing)$ where $P(\varnothing)$ is the probability assigned to the ``no-object" token (see \S \ref{sec:method}). We then filter the boxes with a confidence lower than $0.7$. Finally, we merge the masks corresponding to each of these boxes into one binary mask corresponding to this referring expression. The results are collected in Table \ref{tab:phrasecut}.  Our model is able to produce clean masks for a wide variety of long tailed-concepts covered by PhraseCut. Example predictions from our model on this dataset are given in Appendix \ref{sec:appendix_hyperparameters}. 

\begin{figure*}
 \centering
 \includegraphics[width=\textwidth]{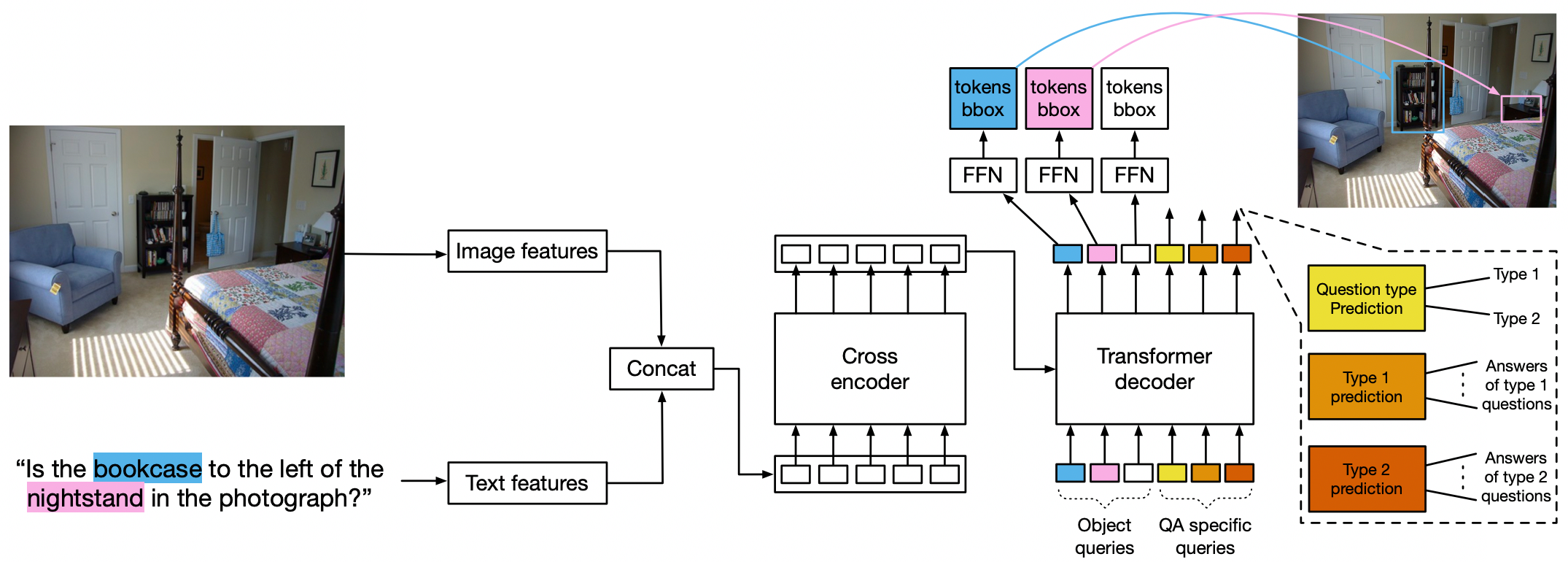}
 \caption{During MDETR pre-training, the model is trained to detect all objects mentioned in the question. To extend it for question answering, we provide QA specific queries in addition to the object queries as input to the transformer decoder. We use specialized heads for different question types.}
 \label{fig:GQA}
\end{figure*}

\textbf{Visual Question Answering} 
We evaluate our hypothesis that modulated detection is a useful component for multi-modal reasoning by fine-tuning our pre-trained model on the GQA dataset. To train MDETR, we use the scene graph provided in GQA to obtain the alignment between question words and the boxes.  Our model architecture is depicted in Fig~\ref{fig:GQA}. Object queries are learned embeddings input to the decoder, each of which can be used to a detect an object. 
Apart from the 100 queries that are used for detection, we use additional queries that specialize in the type of question as well as one that is used to predict the type of question, where the types are defined in the GQA annotations as \textsc{rel}, \textsc{obj}, \textsc{global}, \textsc{cat} and \textsc{attr}. We take our pre-trained model trained for 40 epochs on our combined dataset, and initialise these queries as well as the heads for each of them randomly, and fine-tune first for 125 epochs on the unbalanced \emph{all} GQA split, followed by 10 epochs on the \emph{balanced} split similar to what is done in prior work \cite{li2020oscar, chen2021meta}. During the first 125 epochs, we train the modulated detection losses along with the question answering, but put a weight on question answering loss that encourages the model to focus more on this task. For the balanced split fine-tuning, we only use the question answering loss. 
During inference, the type head predicts the type of question and the answer is taken from that head. Using our model with a Resnet-101 backbone, we not only outperform LXMERT \cite{tan2019lxmert} and VL-T5 \cite{Cho2021UnifyingVT} which use comparable amount of data, but also OSCAR \cite{li2020oscar} which uses magnitude more data in their pre-training. MDETR with the EfficientNet-B5 backbone is able to push performance even higher as reported in Table \ref{tab:GQA}. The NSM model makes use of an external scene graph generation model, while the MMN model makes use of the scene graph and functional programs during training. 
    
\setlength{\tabcolsep}{2pt}
\begin{table}[t]
\begin{center}
\small
\begin{tabular}{cccc}
 \toprule
 Method & Pre-training img data &  Test-dev &  Test-std \\
 \midrule
 MoVie \cite{nguyen_movie_2020} &  -  &  -  & 57.10 \\
 LXMERT\cite{tan2019lxmert}   & VG, COCO (180k)   & 60.0  & 60.33   \\
 VL-T5 \cite{Cho2021UnifyingVT}  & VG, COCO (180k) &  -  & 60.80 \\
  MMN \cite{chen2021meta}  &  -  &  -  & 60.83 \\
 \multirow{2}{*}{OSCAR \cite{li2020oscar}}     & VG, COCO, & \multirow{2}{*}{61.58}  & \multirow{2}{*}{61.62} \\
  & Flickr, SBU (4.3M)        \\
  NSM \cite{Hudson2019LearningBA}   &  -  &  -  & 63.17  \\
  \multirow{3}{*}{VinVL \cite{zhang2021vinvl}}     & VG, COCO, Objects365, SBU & \multirow{2}{*}{65.05}  & \multirow{2}{*}{64.65} \\
  & Flickr30k, CC, VQA,         \\
  & OpenImagesV5 (5.65M)        \\
  \midrule
   MDETR-R101   & VG, COCO, Flickr30k (200k)  & 62.48 &  61.99 \\
  MDETR-ENB5   & VG, COCO, Flickr30k (200k)  & 62.95 & 62.45  \\
 \bottomrule
\end{tabular}
\caption{Visual question answering on the GQA dataset.}
\label{tab:GQA} \vspace{-0.5cm}

\end{center}
\end{table}
\setlength{\tabcolsep}{5pt}

\begin{figure}[t]
        \centering
        \includegraphics[trim={0 0 0 0}, clip, height=2in]{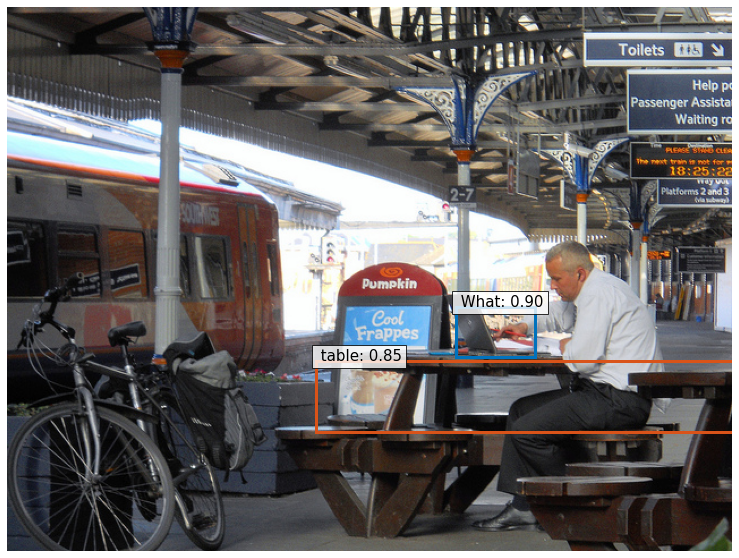}
        \caption{MDETR provides interpretable predictions as seen here. For the question ``What is on the table?", MDETR fine-tuned on GQA predicts boxes for key words in the question, and is able to provide the correct answer as ``laptop". Image from COCO val set. \label{fig:gqa_interpretable}}
    \end{figure}

\subsubsection{Few-shot transfer for long-tailed detection}

\begin{table}[t]
\begin{center}
\small
\begin{tabular}{ccccccc} 
\toprule
Method & Data& AP & AP50 & AP$_\mathrm{r}$ & AP$_\mathrm{c}$ & AP$_\mathrm{f}$   \\
\midrule

Mask R-CNN & 100\% &33.3 & 51.1 & 26.3 & 34.0 & 33.9\\
 
 DETR & 1\% & 4.2 & 7.0 & 1.9 & 1.1 & 7.3 \\
 DETR & 10\% &13.7 & 21.7 &4.1 & 13.2 & 15.9 \\
 DETR & 100\% & 17.8 & 27.5 & 3.2 & 12.9 & 24.8\\
 \midrule
 MDETR & 1\%  & 16.7 & 25.8 & 11.2 & 14.6 &  19.5 \\
 MDETR & 10\% & 24.2 & 38.0 &20.9 & 24.9 & 24.3 \\ 
 MDETR & 100\% & 22.5 & 35.2 & 7.4 & 22.7 & 25.0 \\
\bottomrule
\end{tabular}
\caption{Box AP \emph{fixed} results on LVIS-v1. Since the validation set of LVIS contains some training images from MSCOCO, we report results on the subset of 5k validation images that our model has never seen during training. We call this subset \emph{minival}.
All models use a Resnet 101 as backbone. Mask-RCNN can be regarded as a strong representative of the detection performance of current approaches on this dataset, using bells and whistles such as Repeat Factor Sampling (RFS) to address class imbalance. We use a vanilla DETR pretrained on MSCOCO as a few-shot transfer baseline, and show that our pre-training on natural text improves performance significantly, especially on rare categories.\label{tab:lvisresults}}\vspace{-0.6cm}

\end{center}
\end{table}

Inspired by the success of CLIP~\cite{radford2021learning}, on zero-shot transfer for image classification, we explore the opportunity to construct a useful detector over a given label set from a pre-trained MDETR model. Unlike CLIP, we do not ensure our pre-training dataset contains a balanced representation of all the target classes. By construction, our dataset has no training instances where there are zero boxes aligned to the text, biasing the model to always predict boxes for a given text. This prevents evaluating in a true zero-shot transfer setting,  so we turn instead to a few-shot setting, where the model is trained on a fraction of the available labelled data. We conduct our experiments on the LVIS dataset~\cite{guptaLVISDatasetLarge2019}, a detection dataset with a large vocabulary of 1.2k categories, with a long-tail that contains very few training samples, making it a challenging dataset for current approaches. Federated datasets often pose problems to standard detectors, and require developing specific loss functions~\cite{Tan_2020_CVPR}. However this property makes it well suited to train MDETR: for each positive category, we create a training instance composed of the image and a text version of the class name, and provide as annotations all the instances of this category. For each negative category, we provide the class name and an empty set of annotations.
For inference on a given image, we query each possible class name, then merge the sets of boxes detected on each of the text prompts. This inference scheme costs about 10s/image on a GPU.%

We fine-tune MDETR on three subsets of the LVIS train set, each containing respectively 1\%, 10\% and 100\% of the images. We ensure a balanced sampling of the categories, such that our 1\% set contains at least one positive and one negative examples from each category. We compare to two baselines: the first one is Mask-RCNN trained exclusively on the full training set of LVIS. The other is a DETR model pre-trained on MSCOCO then fine-tuned on the various subsets of the LVIS training set. Our results are shown in Table \ref{tab:lvisresults}.
Following recent recommendation~\cite{daveEvaluatingLargeVocabularyObject2021} on AP evaluation in the context of large vocabulary, we report the box AP \emph{fixed}, obtained by limiting the number of detections per category instead of per image. Even with as little as 1 example per class, MDETR leverages the text pre-training and outperforms a fully fine-tuned DETR on rare categories. We note however that under full fine-tuning on the whole training set, the performance on rare objects drops significantly from 20.9 AP with 10\% data to 7.5 with 100\%, likely due to the extreme class imbalance. We expect that common techniques such as Repeat Factor Sampling will improve the situation in future work.

\section{Related work}
The CLEVR dataset \cite{johnson2017clevr} is a popular vision-language benchmark for reasoning on objects, their relations, and the composition of such relations. A prominent line of work \cite{johnson_inferring_2017,yiNeuralSymbolicVQADisentangling2019,mascharkaTransparencyDesignClosing2018,hu_learning_2017} makes use of the functional programs annotations that are part of the CLEVR dataset. Such approaches tend to dominate on the question answering benchmark, but fail to generalize beyond synthetic data. Conversely, many approaches \cite{perezFiLMVisualReasoning2017, santoro_simple_2017,wangInterpretableVisualReasoning2020,Hudson2018CompositionalAN} learn directly from images or pre-detected objects, with varying amounts of inductive bias tailored to the QA task. Our method can be seen as an in-between: while not explicitly using the program supervision, it is trained to detect objects that are required for performing intermediate reasoning steps.

Recent progress in multi-modal understanding has been mainly powered by pre-training large transformer models to learn generic multi-modal representations from enormous amounts of aligned image-text data \cite{sharma-etal-2018-conceptual}, then fine-tuning them on downstream tasks.
These methods can be divided into single stream \cite{chen2019uniter, li2020oscar, zhang2021vinvl, li2019visualbert} and two-stream \cite{tan2019lxmert, lu2019vilbert, lu202012,su2019vl} architectures depending on whether the text and images are processed by a single combined transformer or two separate transformers followed by some cross attention layers. For both these types, the prevalent approach is to extract visual and textual features independently and then use the attention mechanism of the transformers to learn an alignment between the two. While this approach has improved state of the art results on a wide variety of tasks such as image-text retrieval \cite{zhang2021vinvl}, phrase grounding \cite{li2019visualbert}, image captioning \cite{li2020oscar} and visual question answering \cite{li2021semvlp}, it leaves opportunity for a more tightly knit architecture, such as MDETR, in which information flows between the two modalities at an even earlier stage of the model. Some previous attempts at achieving this using modulated architectures such as \cite{perezFiLMVisualReasoning2017} and \cite{nguyen_movie_2020} show improvements on counting tasks and visual question answering. 

The visual features used by the current state-of-the-art models are extracted using an external pre-trained detector~\cite{anderson2018bottom}, which outputs regions that are noisy, often over-sampled and ambiguous. \cite{li2020oscar} attempts to alleviate the problem of noisy image features by using tags as anchors between the text and images. This is still a weaker form of supervision than in MDETR where we have explicit alignment between words or phrases in text and the objects in the images. To alleviate the constraints implied by fixed vocabulary of concepts, \cite{zhang2021vinvl} trains on a collection of much larger object detection datasets in pursuit of better coverage. \cite{gan2020large} conduct adversarial training on top of existing high performing models pushing performance even higher. Other approaches \cite{yu2020ernie} attempt to incorporate scene graph prediction as part of their pre-training to learn more robust representations. Some recent work also attempts to build multi-purpose multi-modal architectures that are able to tackle a variety of vision-language \cite{Cho2021UnifyingVT} as well as pure language tasks in a single architecture \cite{Hu2021UniTMM}.
A separate line of work that attacks a similar problem to ours but with a much more task specialized model architectures are the single \cite{yang2019fast, chen2019see, li2018referring} and two stage \cite{plummer2020revisiting, hinami2017discriminative} referring expression segmentation and phrase detection models which are designed specifically for this task.

\label{sec:relwork}

\section{Conclusion}
\label{sec:conclusion}
We presented MDETR, a fully differentiable modulated detector. We established its strong performance on multi-modal understanding tasks on a variety of datasets, and demonstrated its potential in other downstream applications such as few-shot detection and visual question answering. We hope that this work opens up new opportunities to develop fully integrated multi-modal architectures, without relying on black-box object detectors.

\section*{Acknowledgements}
We would like to thank Kyunghyun Cho, Ethan Perez, Sergey Zagoruyko and Francisco Massa for helpful discussions and feedback at various points of this project. We would also like to thank Alex Kirillov and Ross Girshick for their help with the LVIS evaluations, Justin Johnson for test set evaluation on CLEVR, Bryan Plummer for discussions on best evaluation practices for phrase grounding and finally Runtao Liu and Chenxi Liu for their feedback on dataset construction and evaluation for CLEVR referring expressions. 

Aishwarya Kamath was supported in part by AFOSR award FA9550-19-1-0343 and Nicolas Carion by a grant from NVIDIA. 

{\small
\bibliographystyle{ieee_fullname}
\bibliography{main}
}

\clearpage
\appendix\newpage
\onecolumn

\section{Model details and hyperparameters}
\label{sec:appendix_hyperparameters}

\paragraph{Pre-training hyperparameters}
MDETR follows the pre-train then fine-tune strategy by first training on our constructed combined dataset for 40 epochs followed by fine-tuning on the respective downstream task. We train our model using AdamW~\cite{loshchilovDecoupledWeightDecay2019}, a variant of Adam~\cite{kingmaAdamMethodStochastic2017} better suited for weight decay. We use exponential moving average (EMA) with a decay rate of 0.9998, and a weight-decay of $1e^{-4}$. The backbone and the transformer have a constant learning rate of respectively $1e^{-4}$ and $1e^{-5}$ for 35 epochs, after which their learning rate is reduced by a factor of 10. For the language model's learning rate, we use a linear decay with warmup schedule, increasing linearly to $5e^{-5}$ during the first 1\% of the total number of steps, then decreasing linearly back to 0 for the rest of the training.
\begin{figure*}[h]
 \centering
 \includegraphics[width=0.82\textwidth]{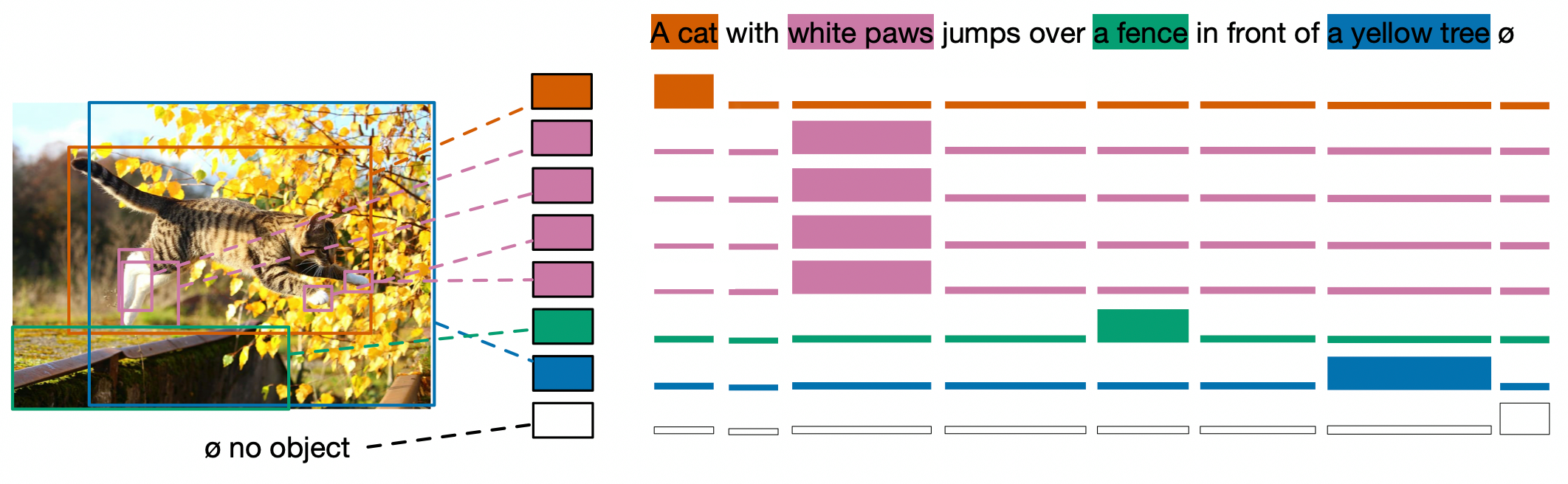}
 \caption{Illustration of the soft-token classification loss. For each object, the model predicts a distribution over the token positions in the input sequence. The weight of the distribution should be equally spread over all the tokens that refer to the predicted box.}
 \label{fig:soft_loss}
\end{figure*}

\paragraph{Flickr30k} Our results on Flickr30k are evaluated using the pre-trained model, without any additional fine-tuning as we found that it brings no additional gains.
For evaluation, we must rank the boxes associated with each phrase. Since there might be several phrase in the same sentence, we must provide a ranking for each and every such phrase. To that end, we use the prediction from the soft-token classification loss. In the example depicted in Fig \ref{fig:soft_loss}: to rank the boxes for the phrase ``a cat", we use the probability mass that each query assigns to the positions that correspond to ``a cat" in the sentence ``a cat with white paws jumps over a fence in front of a yellow tree" (in this example, the first few tokens). Through this approach,  the red box is found to be the highest-ranked box. On the other hand, if we want the boxes corresponding to ``the fence", we sort them according to the corresponding token positions, and in this case we find the green box as the top-scoring one.

\begin{figure*}[h]
    \centering
    \begin{subfigure}[t]{0.2\textwidth}
        \centering
        \includegraphics[height=2.5in]{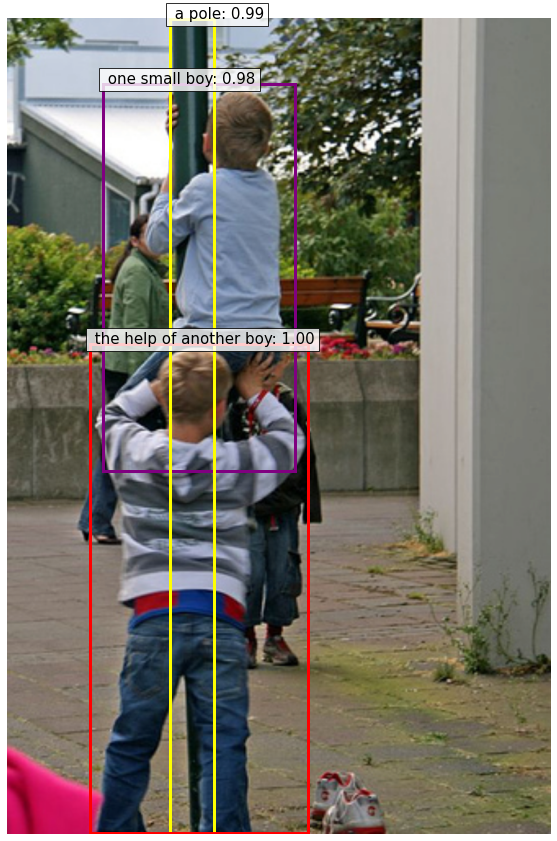}
        \caption{``one small boy climbing a pole with the help of another boy on the ground"}
    \end{subfigure}%
    \hspace{2mm}
    \begin{subfigure}[t]{0.3\textwidth}
        \centering
        \includegraphics[height=2.5in]{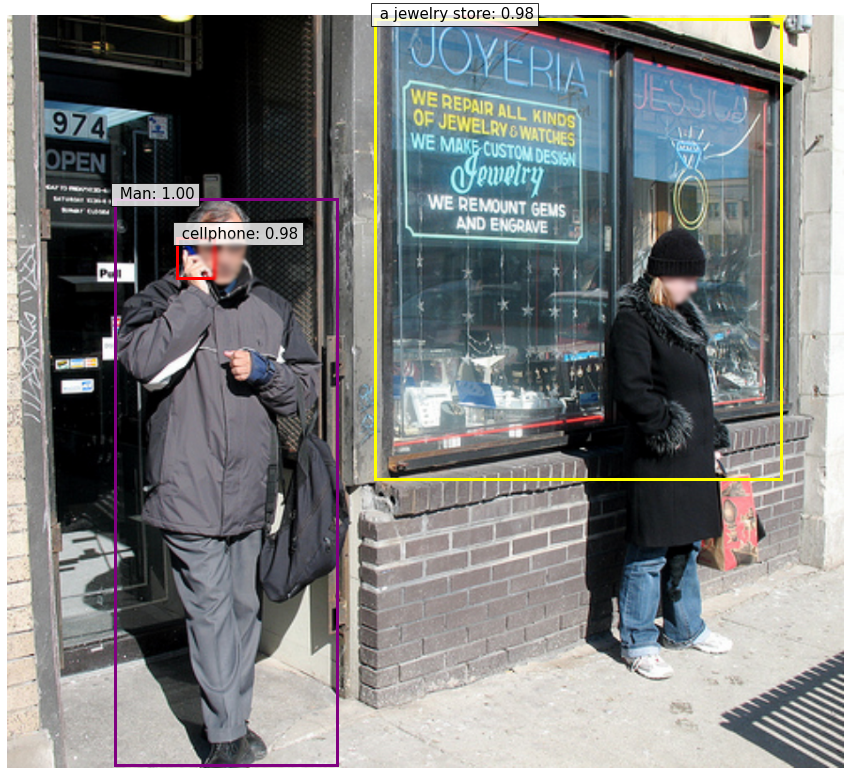}
        \caption{``A man talking on his cellphone next to a jewelry store"}
    \end{subfigure} \\
    \begin{subfigure}[t]{\textwidth}
        \centering
        \includegraphics[height=2in]{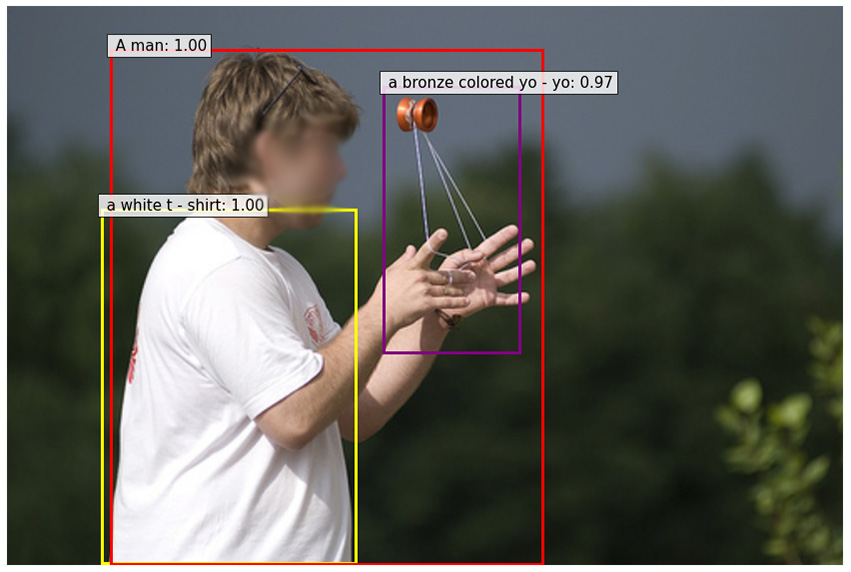}
        \caption{``A man in a white t-shirt does a trick with a bronze colored yo-yo"}
    \end{subfigure}
    \caption{Examples of phrase grounding on the Flickr30k dataset}
    \label{fig:flickr_examples}
\end{figure*}

\begin{figure*}[t!]
    \centering
    \begin{subfigure}[t]{0.33\textwidth}
        \centering
        \includegraphics[width=\textwidth]{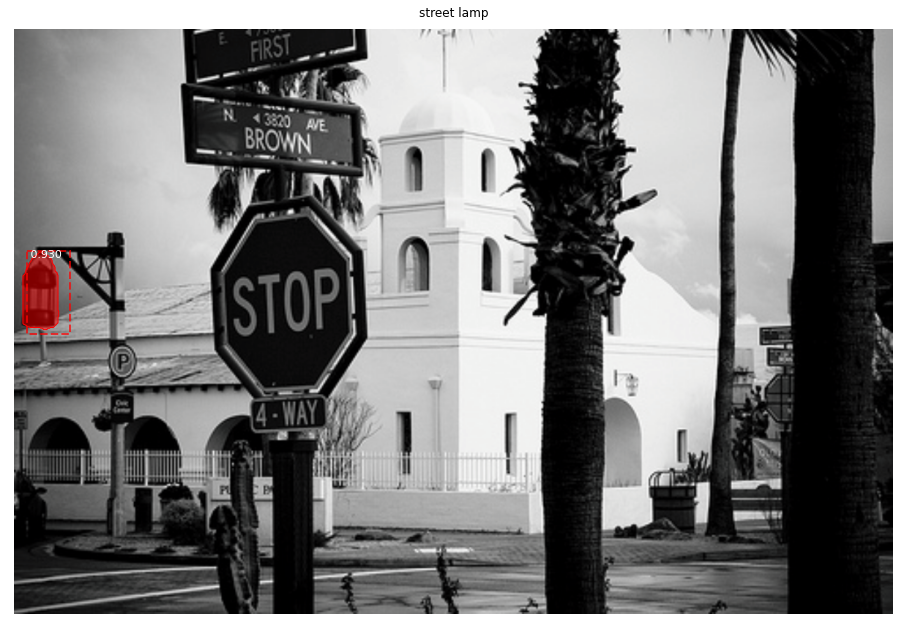}
        \caption{Query: ``street lamp"}
    \end{subfigure}%
    \begin{subfigure}[t]{0.33\textwidth}
        \centering
        \includegraphics[width=\textwidth]{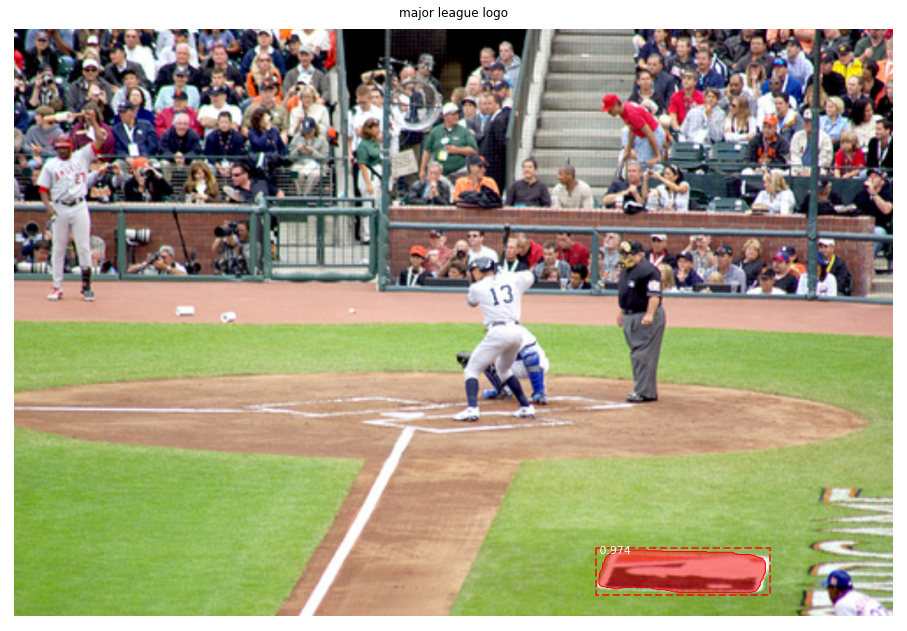}
        \caption{Query: ``major league logo"}
    \end{subfigure} 
    \begin{subfigure}[t]{0.33\textwidth}
        \centering
        \includegraphics[width=\textwidth]{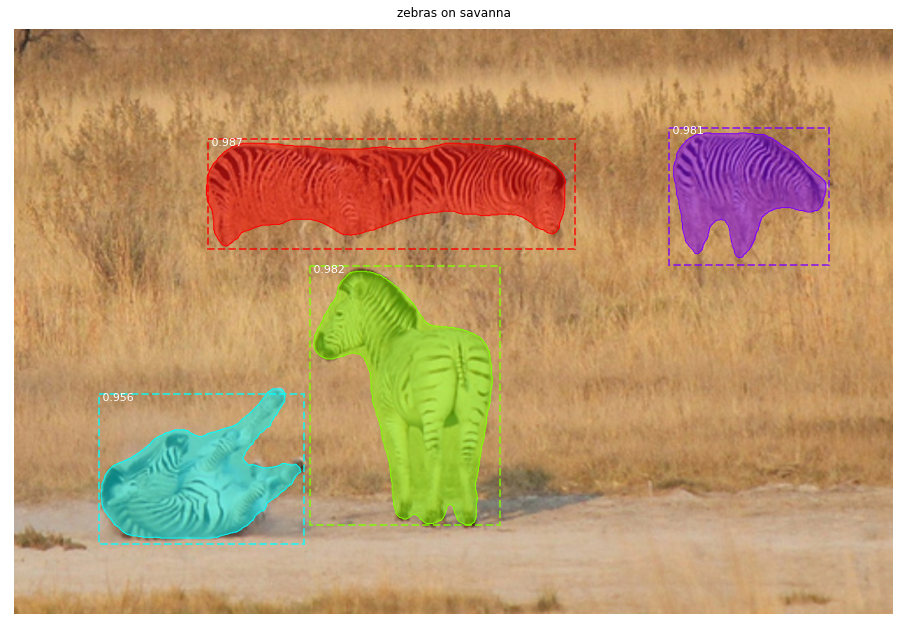}
        \caption{Query: ``zebras on savanna"}
    \end{subfigure}
    \caption{Qualitative segmentation examples on the phrasecut dataset
    \label{fig:phrasecut_examples}}
\end{figure*}

\begin{figure*}[t!]
    \centering
    \begin{subfigure}[t]{0.31\textwidth}
        \centering
        \includegraphics[width=\textwidth]{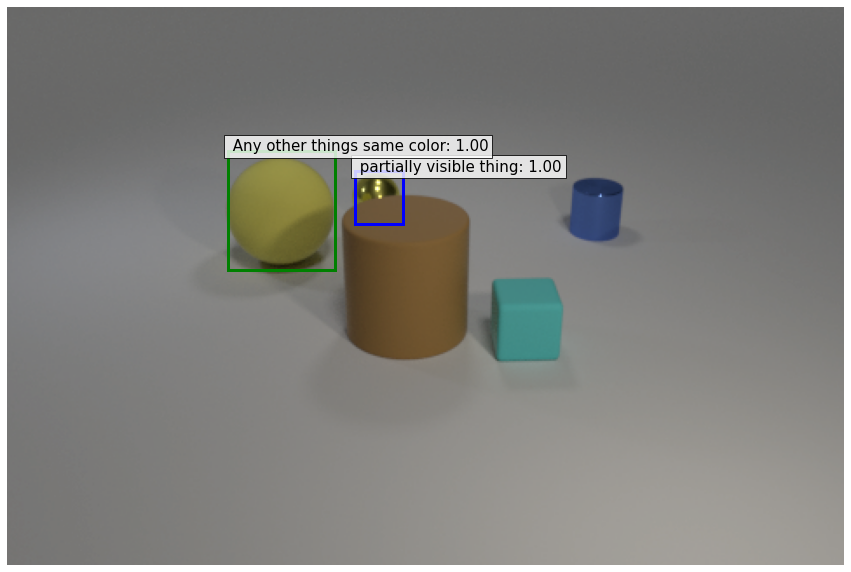}
        \caption{Query: ``Any other things that are the same color as the partially visible thing(s)"}
    \end{subfigure}%
    \begin{subfigure}[t]{0.31\textwidth}
        \centering
        \includegraphics[width=\textwidth]{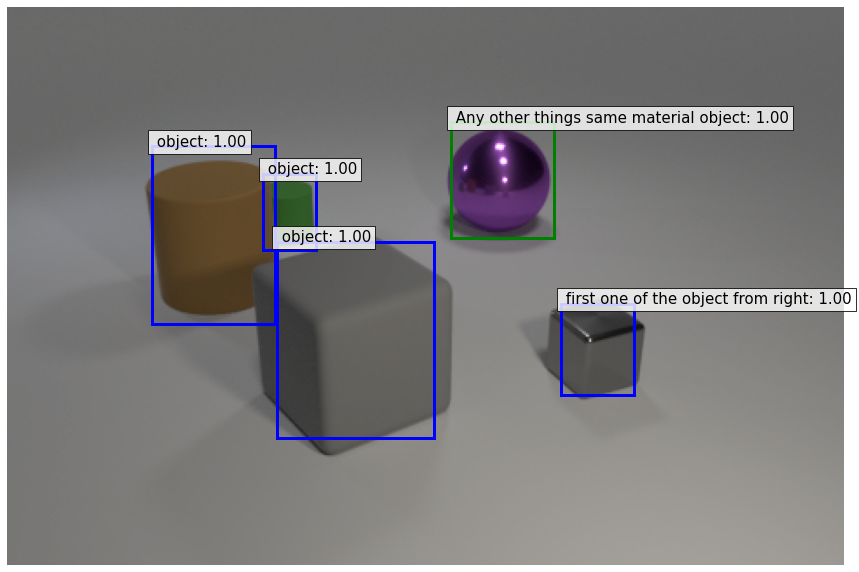}
        \caption{Query: ``Any other things that are the same material as the first one of the object(s) from right"}
    \end{subfigure} 
    \begin{subfigure}[t]{0.31\textwidth}
        \centering
        \includegraphics[width=\textwidth]{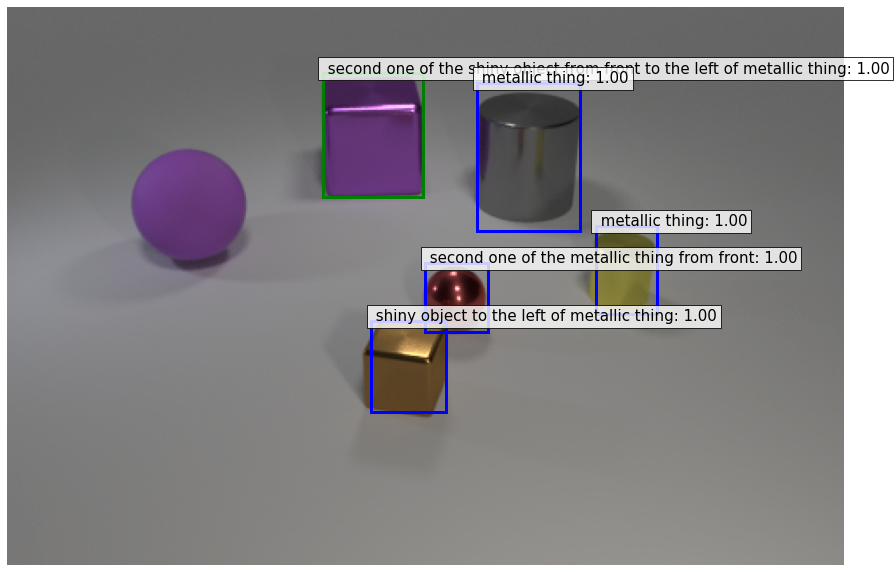}
        \caption{Query: ``The second one of the shiny object(s) from front that are to the left of the second one of the metallic thing(s) from front"}
    \end{subfigure}
    \caption{Qualitative example from the CLEVR-REF+ dataset. When the model predicts a box that is referred to, we display it in green. The other boxes are intermediate reasoning steps and are depicted in blue.
    \label{fig:phrasecut_examples}}
\end{figure*}

\begin{figure*}[t!]
    \centering
    \begin{subfigure}[t]{0.33\textwidth}
        \centering
        \includegraphics[width=40mm,]{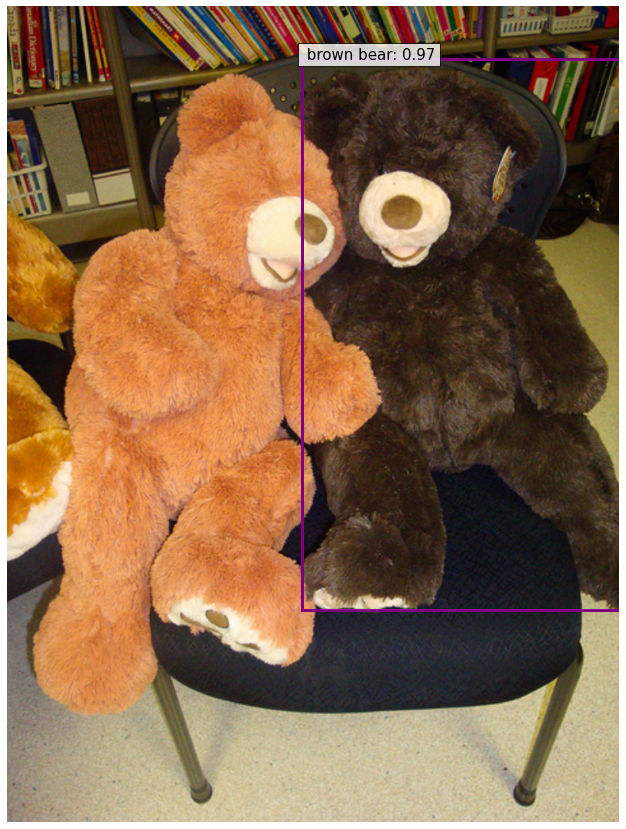}
        \caption{``brown bear"}
    \end{subfigure}%
    \hspace{-3mm}
    \begin{subfigure}[t]{0.33\textwidth}
        \centering
        \includegraphics[width=60mm]{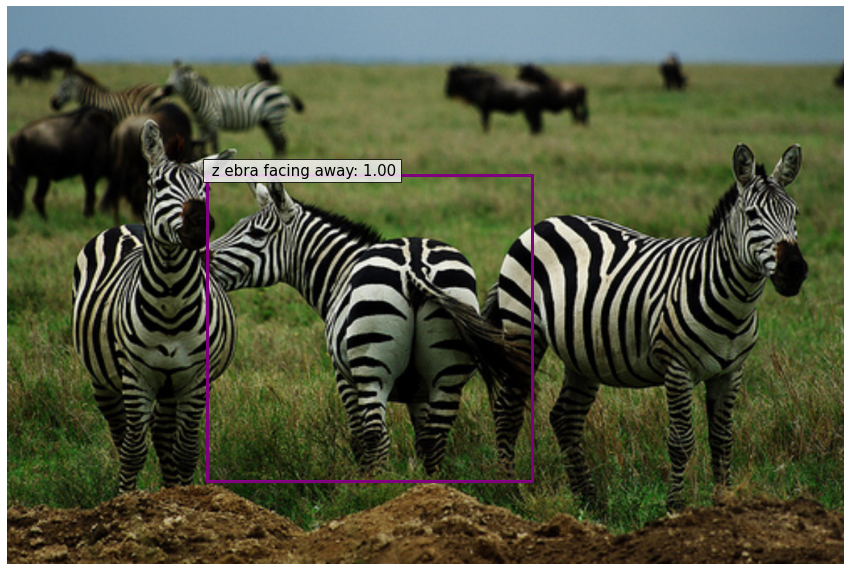}
        \caption{``zebra facing away"}
    \end{subfigure} \\
    \begin{subfigure}[t]{0.3\columnwidth}
        \centering
        \includegraphics[width=\textwidth]{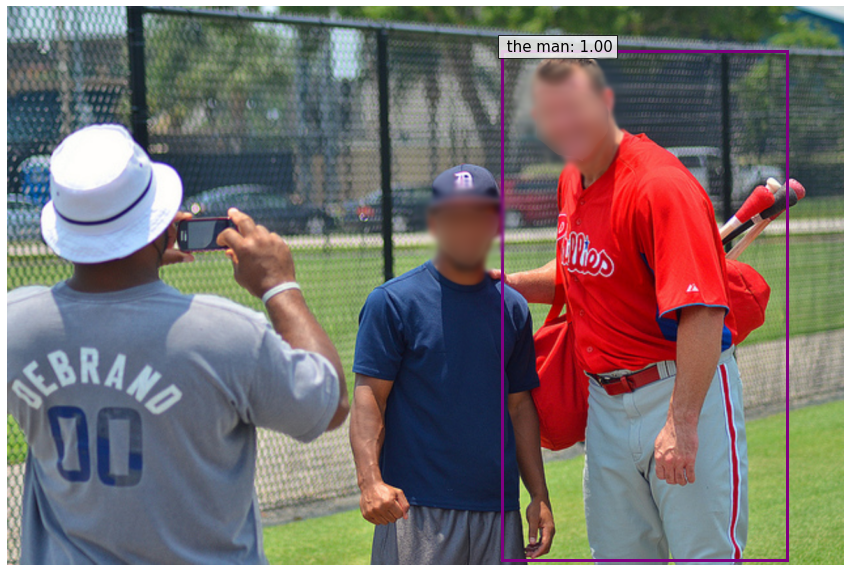}
        \caption{``the man in the red shirt carrying baseball bats"}
    \end{subfigure}
    \hspace{2mm}
    \begin{subfigure}[t]{0.3\textwidth}
        \centering
        \includegraphics[width=\textwidth,scale=0.5]{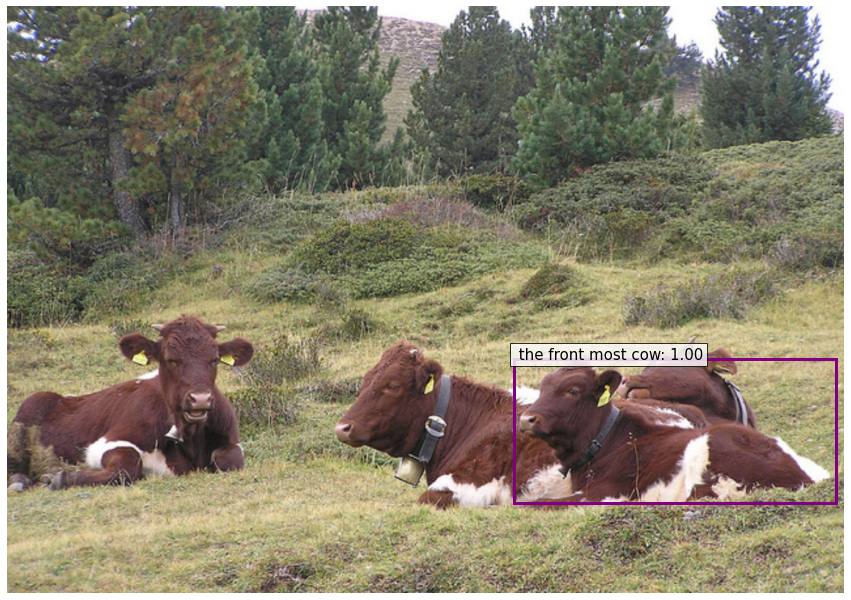}
        \caption{``the front most cow to the right of the other cows"}
    \end{subfigure}%
    \caption{Examples from RefCOCO, RefCOCO+ and RefCOCOg datasets. Fig(a) taken from RefCOCO, Fig(b) from RefCOCO+ and Fig(c) and (d) are taken from RefCOCOg, in which the expressions are much longer on average and contain more descriptive language than in RefCOCO and RefCOCO+. Even when the expressions are long, we train our model to align the box to the root of the phrase, for eg. ``the man" in (c). The model however, still has access to the whole text and uses it to disambiguate between the two men in the image.
    \label{fig:refexp_examples}}
\end{figure*}

\paragraph{Referring Expression Comprehension} For the Referring Expression Comprehension task, there is a stark difference in how the data is presented to the model in terms of density of annotation. For all other datasets that we use in our pre-training, each noun phrase in the sentence is annotated with its respective box, if available. On the other hand, in RE comprehension, the task is to align the whole referring expression with the corresponding box, possibly by needing to disambiguate between different occurrences of the same category of object. In other words, our model now needs to predict one box per expression. A problem with this setting is that our box-token contrastive alignment as well as soft token prediction losses get very diluted signal if we align the whole sentence to the box. To alleviate this, we pre-process the text using Spacy \cite{spacy} to extract the root of the sentence using a dependency parser. The tokens from this root phrase are used to align to the box.
Fine-tuning on this dataset is therefore crucial for good performance. We fine-tune for 5 epochs on the RefCOCO, RefCOCO+ and RefCOCOg datasets with a learning rate of $1e^{-5}$ for the backbone and $5e^{-5}$ for the rest of the network. We use a learning rate drop by a factor of 10 after 3 epochs. For the text encoder we use a learning rate of $1e^{-5}$, with a linear decay with warmup schedule, warming up over the first 1\% of steps and then decaying to 0 linearly. At inference time, to detect a given expression, we feed it to the model alongside the image. We then rank the 100 detected boxes according to the probability that the box corresponds to an actual object (as opposed to a ``no object"). If $P(\varnothing)$ is the probability mass assigned to the ``no object" label, then we rank by increasing order of $P(\varnothing)$, or equivalently by decreasing order of $1-P(\varnothing)$. We show an example in Fig \ref{fig:refexp_examples} of the box predicted by our model for the corresponding referring expressions. 
In addition there is quite some variety in the type of text annotations from the three datasets. Both RefCOCO and RefCOCO+ were collected in a timed game setting whereas RefCOCOg was not. This led to differences in the length and diversity of language used in the different datasets. RefCOCO+ disallowed usage of location words to describe objects or disambiguate between multiple occurrences of the same object, focusing more on appearance based descriptions. RefCOCOg consists of expressions more than twice the length (on average) of the others and with more flowery and descriptive language.

We also fine-tuned the EfficientNetb5 model on these datasets but did not see much improvement over the EfficientNetB3 model, and we believe this is due to the smaller size of these datasets causing the larger model to overfit. 

\paragraph{PhraseCut}
Detection: For this phase, we use a batch size of 64, a learning rate of $1e^{-5}$ for the text encoder and backbone and $5e^{-5}$ for the rest of the network, and exponential moving average (EMA) of the network weights with a decay of 0.9998. segmentation: For this stage, we use a lr of $5e^{-4}$ and no EMA. See \cite{carion2020end} for additional details.

\section{CLEVR Experiments}
\label{appendix_section:CLEVR_results_detailed}
\subsection{Dataset details}
\label{sec:clevr_dataset_details}
The CLEVR dataset consists of 3D-rendered scenes containing between 3 and 10 objects of various shapes, material, size and color. Each of these scenes is associated with about 10 questions that are formulated about the visible objects, generated from a fixed set of templates. Each question is guaranteed to be answerable, and the annotations further provide a functional program that describe how to compute the answer using elementary reasoning steps. The total training set contains 70k images and slightly less than 700k questions. Overall, the visual aspect of this task, \emph{ie} the scene parsing, is not really challenging by modern standards, since the set of objects is limited, unambiguous, and there are no visual distractors. The only challenging cases occur in the event of heavy occlusion, where it might be hard to make out the shape of the occluded object, or in some cases where the question requires comparing ambiguous spatial relations (eg. asking which object is the closest to the camera in a setting where they are visually nearly tied). On the other hand, the text understanding aspect is more involved, since the questions can be quite complex, involving up to 20 reasoning steps.
Unlike several successful approaches to CLEVR~\cite{johnson_inferring_2017, Hudson2018CompositionalAN, mascharkaTransparencyDesignClosing2018, yiNeuralSymbolicVQADisentangling2019} , MDETR doesn't incorporate any special inductive bias to cope with such complex reasoning tasks. In this section, we show that despite its relatively straight-forward formulation, our approach competes with state-of-the-art models on the question answering task.

The first ingredient required for training MDETR is bounding box annotations for objects in the image. The original CLEVR dataset doesn't provide any, so we use the scene graphs from the dataset to re-create the original scene in the 3D-renderer Blender, then use some of its functionalities to extract the segmentation masks of the visible parts of the objects, and deduce the bounding boxes from that. The main complication is that the original rendering involved some non-deterministic jittering of the camera's position and rotation, leading to some potential discrepancies in the computed boxes. To minimize the error, we use the known 3D position of each object, as well as their known 2D location in the rendered image to optimize the camera parameters using a gradient-based approach. The final boxes obtained using this approach are accurate within a 10-pixel error margin, which we deem appropriate for our purposes.

The second ingredient required is the alignment between bounding boxes and tokens in the question. MDETR is trained to predict \textit{only} objects that are referred to in the question. For example, in the question ``What is the color of the cube in front of the small cylinder?", we provide an annotation for both the small cylinder (an intermediate step) and the cube (the main subject), and none of the other objects present in the scene. We use the functional programs that are part of the original CLEVR annotations to extract this set of objects, along with their corresponding text tokens in the original question.

\setlength{\tabcolsep}{4pt}
\begin{table*}[t]
\begin{center}
\small
 \begin{tabular}{cccccccccccccc} 
 \toprule
Method & \multicolumn{6}{c}{CLEVR} & \multicolumn{2}{c}{CLEVR-Humans} & \multicolumn{2}{c}{CoGenT} &CLEVR-Ref+ \\ [0.5ex]
        & Overall & Count & Exist & Comp. Num & Query     & Comp. Att   & Before FT & After FT & TestA & TestB& Acc  \\
 \midrule 
 MAttNet\cite{yu_mattnet_2018}&-&-&-&-&-&-&-&-&-&-&60.9\\
 MGA-Net\cite{Zheng_2020_ACCV}&-&-&-&-&-&-&-&-&-&-&80.1\\
 FiLM\cite{perezFiLMVisualReasoning2017} &97.7&94.3&99.1&96.8&99.1&99.1 &56.6 & 75.9 & 98.3 & \textbf{78.8}&-\\

 MAC \cite{Hudson2018CompositionalAN}   & 98.9    & 97.1 & 99.5  & 99.1 & 99.5 & 99.5 & 57.4 & 81.5 & - & - &-\\
 NS-VQA\cite{yiNeuralSymbolicVQADisentangling2019}$^*$ & \textbf{99.8} & \textbf{99.7} & \textbf{99.9} & \textbf{99.8} & 99.8 & 99.8 & -&67.8&\textbf{99.8} & 63.9 &-\\
 OCCAM \cite{wangInterpretableVisualReasoning2020} & 99.4 & 98.1 & 99.8 & 99.0 & 99.9 & 99.9 & - & - & - & - &- \\
 MDETR & 99.7 & 99.3 & \textbf{99.9} & 99.4 & \textbf{99.9} & \textbf{99.9} & \textbf{59.9} & \textbf{81.7} & \textbf{99.8} & 76.7 &\textbf{100}\\
\bottomrule
\end{tabular}
\caption{Results on CLEVR-based datasets. We report accuracies on the test set of CLEVR, including the detail by question type. On CLEVR-Humans, we report accuracy on the test set before and after fine-tuning. On CoGenT, we report performance when the model is trained in condition A, without finetuning on condition B. On CLEVR-Ref+, we report the accuracy on the subset where the referred object is unique. *indicates method uses external program annotations}
\label{tab:clevrresults}
\end{center}
\end{table*}

\begin{figure*}
 \centering
 \includegraphics[width=\textwidth]{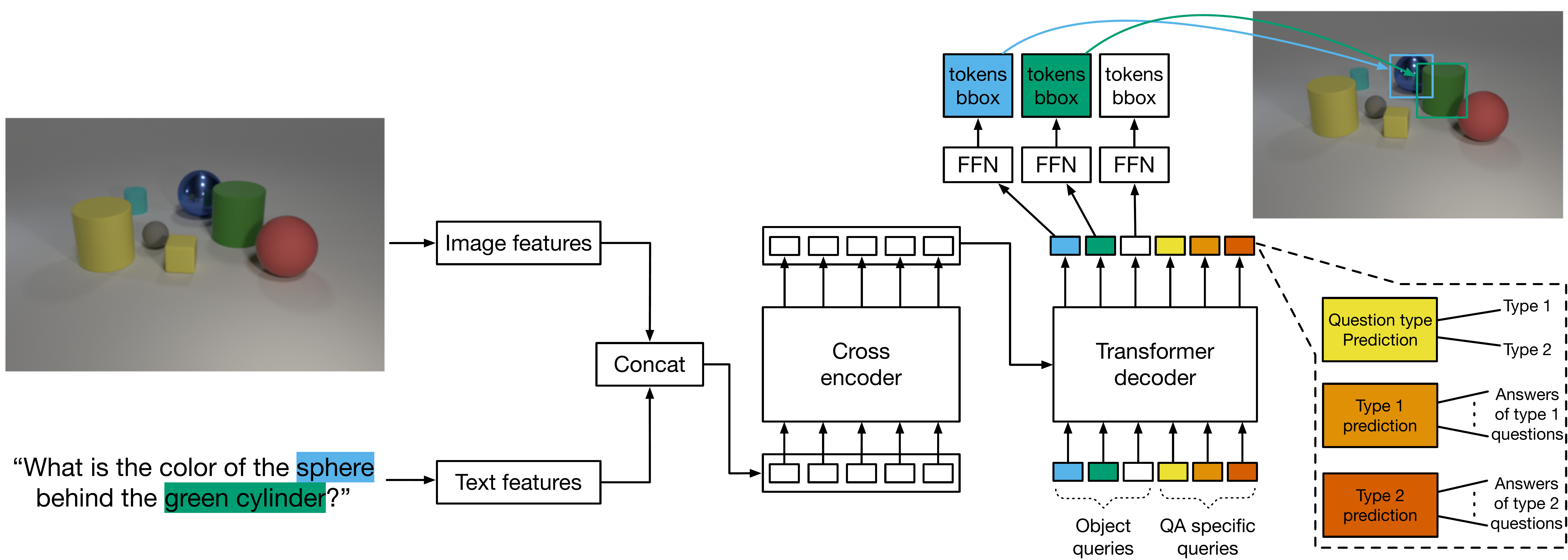}
 \caption{During MDETR pre-training, the model is trained to detect all objects mentioned in the question. To extend it for question answering, we provide QA specific queries in addition to the object queries as input to the transformer decoder. We use specialized heads for different question types.}
 \label{fig:QA}
\end{figure*} 

\subsection{Training details}
\label{sec:clevr_training_details}
\textbf{Model} We use a ResNet-18~\cite{he2016deep} from Torchvision, pre-trained on ImageNet~\cite{ILSVRC15} as the convolutional backbone. For the text-encoder, we use a pre-trained DistilRoberta~\cite{Sanh2019DistilBERTAD} from HuggingFace \cite{Wolf2020TransformersSN}.
The final transformer is the same as DETR, with 6 encoder layers as well as 6 decoder layers, and 8 attention heads in the attention layers. We reduce the number of object queries to 25, since the maximum number of objects to be detected is low.

\textbf{Pre-training} We first train the model only on the modulated detection objective, on our CLEVR-Medium subset, for 30 epochs. Following DETR training procedure, the transformer and the backbone use a learning rate of respectively $1e^{-4}$ and $1e^{-5}$, and we reduce them by a factor of 10 at the 20th epoch. The text encoder uses a linear decay with warmup schedule, with a warm-up to $5e^{-5}$ over the first 1\% of the training steps. 

\textbf{QA-finetuning} For question answering, we take our pre-trained checkpoint, add the (untrained) question queries and their corresponding heads, then train on the full CLEVR dataset for 30 epochs, following the exact same learning rate schedule, with both the modulated detection as well as question answering losses.  As depicted in Fig.12, we use additional queries in the transformer decoder to answer each type of question in CLEVR: numerical, binary and attributes. We supervise each of these heads using a standard cross-entropy loss. We monitor the accuracy on the validation set to apply early-stopping. Finally, for CLEVR-Humans, we further fine-tune for 60 epochs, with the learning-rate drop occuring at epoch 40.

\subsection{Results and discussion}
The results are collected in Table \ref{tab:clevrresults}. On CLEVR, we closely match the performance of NS-VQA\cite{yiNeuralSymbolicVQADisentangling2019}, a method that uses external supervision in the form of ground-truth program annotations, and clearly surpass the performance of methods which like us, don't use this extra supervision signal. We then evaluate the generalization capability of our model. 

\textbf{CLEVR-Humans} \cite{johnson_inferring_2017} is a dataset of human-generated questions on CLEVR images. It tests the robustness of the model to new vocabulary and and different reasoning primitives. In a zero-shot setting, we improve substantially over the best competing model. We credit this improvement to our pre-trained language model. After fine-tuning, the gap narrows, suggesting that additional developments may be required to further improve performance. 

\textbf{CoGenT} is a test for compositional generalization. The evaluation protocol consists in training on a set A, where the spheres can be any color but the cubes are either gray, blue, brown or yellow, and the cylinders are red,
green, purple or cyan. We then evaluate in a zero-shot manner on a split B which has the opposite color-shape pairs for cubes and cylinders. Similar to other models, we observe a significant generalization gap. On closer inspection, the biggest drop in accuracy occurs on questions querying the shape of an object (from 99.98\% on testA to 34.68\% on testB), suggesting that the model has learnt strong spurious biases between shape and color. 

\textbf{CLEVR-REF+} Finally, we evaluate our model on CLEVR-REF+\cite{liuCLEVRRefDiagnosingVisual2019}, a referring expression comprehension dataset built on CLEVR images. For each object query, we train an additional binary head to predict whether or not the query corresponds to an object being referred to (as opposed to an auxiliary object in the sentence, that we detect as well). Following \cite{liuCLEVRRefDiagnosingVisual2019}, we evaluate accuracy on the subset of expressions that refer to a unique object, measured as whether the top ranked box has an IoU of at least 0.5 with the target box. Using the aforementioned binary prediction to rank the boxes, our model correctly ranks in first position a valid box for each of the examples of the validation set, leading to an accuracy of 100\%, greatly outperforming prior work.

\subsection{Ablations}
\label{sec:clevr_ablations}
We use CLEVR as a test bed to ablate several aspects of our model. Depending on the ablation, we report either the accuracy on the question answering task on the validation set of CLEVR, and/or the detection performance on this dataset, measured as a class-agnostic Average Precision (AP). When inspecting the modulated detection capabilities of the model, we use class agnostic Average Precision (AP) to evaluate the model. In the unconditional detection case, DETR is able to detect all boxes perfectly. When evaluated on the task of modulated detection, the AP metric therefore captures the model's capability for text understanding since now only the boxes relevant to the query must be detected. 
To put this in context, a model that detects all boxes even when given a text query (thereby ignoring the text completely) gets an AP of around 60. The goal is to achieve an AP close to 100 which would imply the model only finds the relevant boxes. 

\subsubsection{Loss ablations}
We first ablate the various parts of our loss. The results are summarized in Table~\ref{tab:clevr_abl_med} We report modulated detection results on the CLEVR-Medium subset, that we constructed by removing data-points from CLEVR where the same object is referred to by distinct parts of the question. In these ablations, we consider only the performance of the detector, and not the question answering capability. As a result, the queries related to question answering are not present and we do not propagate any QA related loss.

\textbf{Contrastive loss} We first ablate the impact of the contrastive loss, by training a model without it. In this situation, the alignment must occur solely through the soft-token classification loss. 
As shown in Table~\ref{tab:clevr_abl_med}, removing this loss results in a drastic drop in AP. More specifically, when evaluating the model, it becomes apparent that it is able to filter the objects based on some attributes (in particular their shape and size) but not others (in particular color and texture). It is unclear what drives the model in this local sub-optima, nor what statistical shortcut it is leveraging to correctly identify shapes and sizes. However, it shows that solely predicting the spans of the text query associated with each object is not sufficient to learn proper alignment. The contrastive loss, which forces object-queries to be similar to their corresponding text-token, is thus necessary.

\textbf{Soft-token classification loss}
We now study whether predicting the text-spans associated with each object is necessary, provided that we propagate the contrastive loss. Instead of predicting a distribution over span, we construct a simplified version of MDETR which only predicts a binary label for each object query: ``object" or ``no-object" ($\varnothing$). This formulation is equivalent to the vanilla DETR classification loss, with one object class. We observe similar results as the previous ablation, namely a sharp decline in AP and a model that only understands half of the attributes correctly. We thus conclude that both ingredients of our loss are indeed required.

\subsubsection{Question answering ablations}

Finally in Table~\ref{tab:clevr_abl_hard} we ablate two aspects of our training recipe that differ with previous approaches:
\begin{itemize}
    \item \textbf{Curriculum}: We evaluate a model trained directly on the full CLEVR training set, without our modulated detection pretraining on CLEVR-medium. Similarly as the previous section, the model learns to detect only a subset of attributes, leading to poor QA accuracy.
    \item \textbf{Single QA head} In our approach, we train a specialized head for each type of question (numerical, binary, or categorical over the attributes). This differs from previous approaches that usually cast it as a single classification over all possible answers. As shown in  Table~\ref{tab:clevr_abl_hard}, this separation has a big impact on the final accuracy. We hypothesize that it enables the attention pattern for each question type to specialize accordingly to the task, there-by yielding better performance. 
\end{itemize}

\begin{table}[t]
\begin{center}
\small
\begin{tabular}{lc}
 \toprule
 Model & AP \\
 \midrule
Baseline & 99.0\\
\ - contrastive loss& 83.2\\
\ - soft-token classification & 87.7\\
 \bottomrule
\end{tabular}
\caption{Ablation results on modulated detection on CLEVR-Medium. We report the class-agnostic AP. See text for details.
\label{tab:clevr_abl_med}}
\end{center}
\end{table}
\begin{table}[t]
\begin{center}
\small
\begin{tabular}{lcc}
 \toprule
 Model & Detection AP & QA accuracy\\
 \midrule
Baseline & 99.0 & 99.7\\
No curriculum & 89.7&68.2\\
Single QA head & 99.0 & 90.1\\
 \bottomrule
\end{tabular}
\caption{Ablation results on the validation set of CLEVR. We report the class-agnostic AP and the question answering accuracy. See text for details.
\label{tab:clevr_abl_hard}}
\end{center}
\end{table}

\begin{figure}[t]
     \centering
     \begin{subfigure}[b]{0.48\textwidth}
         \centering
         \includegraphics[width=\textwidth]{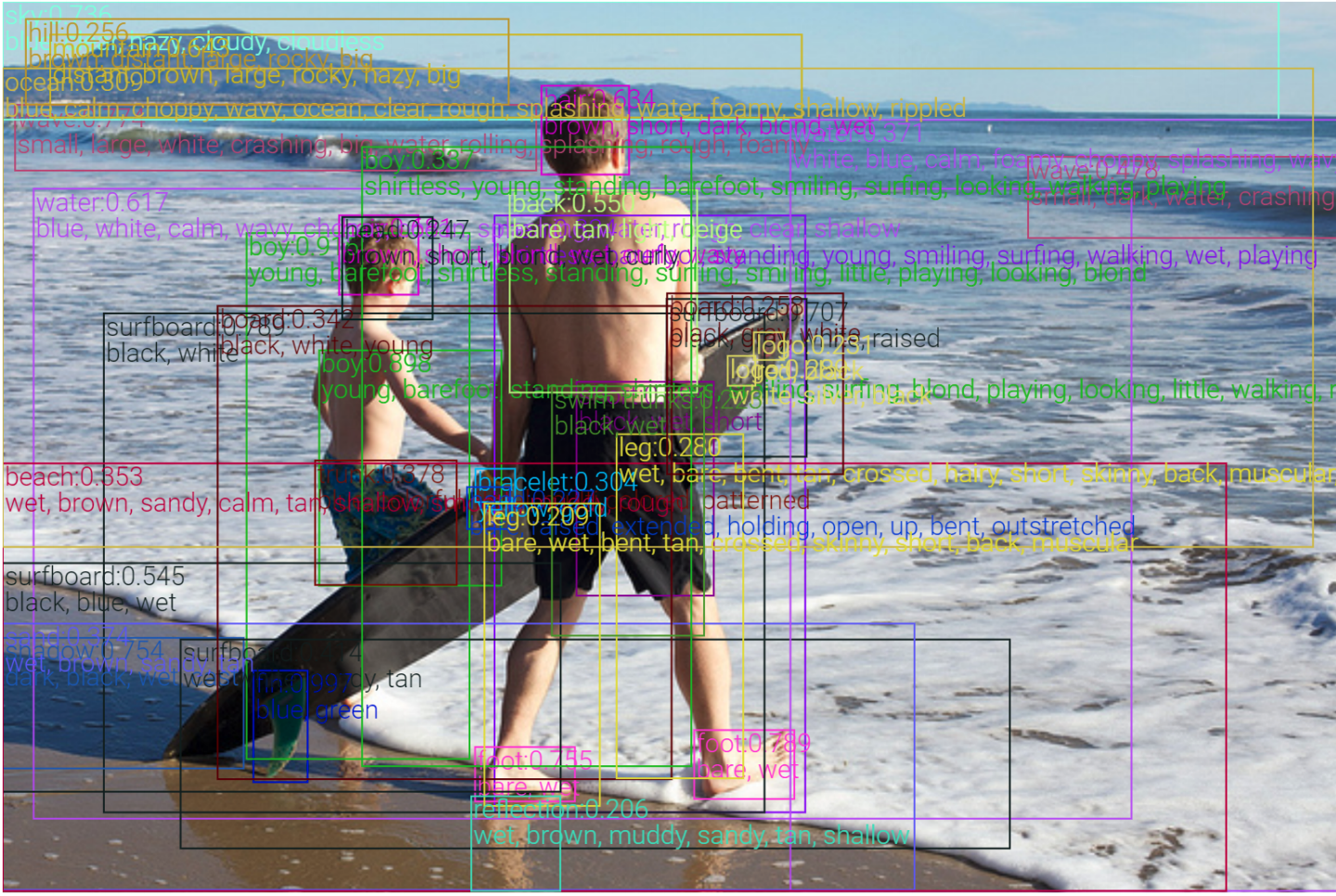}
         \caption{Current object detection pipeline outputs, predicting all possible objects in the image. This extensive annotation is essential to multi-modal understanding systems that treat detection as a black box.\label{fig:main_idea_1}}
     \end{subfigure}
     ~
     \begin{subfigure}[b]{0.48\textwidth}
         \centering
         \includegraphics[width=\textwidth]{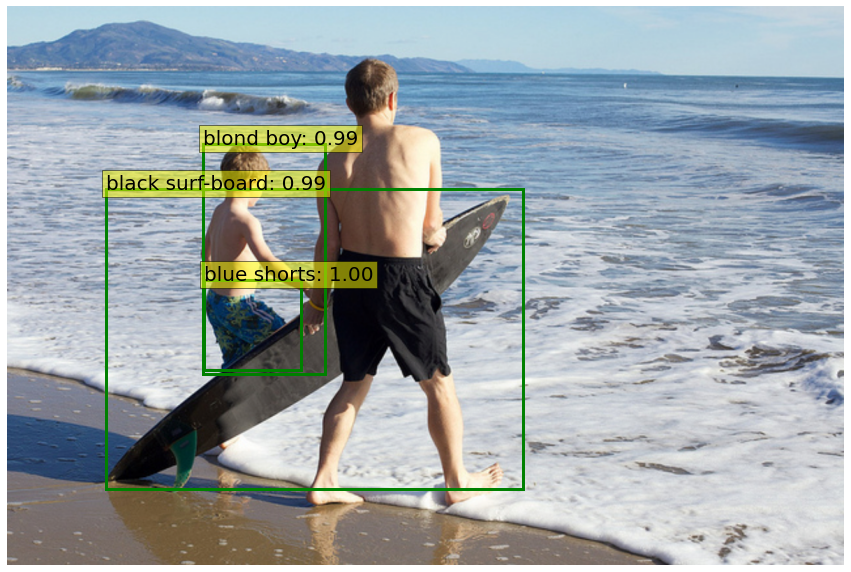}
         \caption{ \Alg predicts boxes relevant to the caption and labels them with the corresponding spans from the text. 
         Here we use the caption: ``blond boy wearing blue shorts. a black surf-board."\label{fig:main_idea_2}}
         
     \end{subfigure}
        \caption{Modulated detection using MDETR vs detection output for a current state-of-the-art multi-modal understanding system. Image taken from \cite{zhang2021vinvl}\label{fig:three graphs}}
\end{figure}

\section{Dataset constructions}
\label{sec:appendix_datasets}
\textbf{MS COCO} On the COCO dataset, we include annotations from the referring expressions datasets (RefCOCO \cite{yu_modeling_2016}, RefCOCO+ \cite{yu_modeling_2016} and RefCOCOg \cite{mao_generation_2016} datasets). 
By construction, in this dataset, each referring expression is a whole sentence that describes one object in the image, where the constituent noun phrases from the sentences are not themselves annotated. For example, in Figure \ref{fig:refexp}, the caption would be ``the person in  the grey shirt with a watch on their wrist.", where only the person would be annotated and not the grey shirt or their watch. To avoid ambiguity, we perform some text pre-processing using SpaCy \cite{spacy} to extract the \textit{root} of the referring expression. This is used in our soft token prediction as well as the contrastive alignment loss for aligning to the referred box. The auxiliary objects (in this example the shirt and the watch) are ignored altogether.

\textbf{Visual-Genome}
We use annotations from VG regions, a dataset having diverse descriptions of a wide variety of objects, often having a very high degree of descriptive detail and covering several concepts. 
By construction, the VG dataset comprises a lot of redundant annotations. We detect redundant sentences by normalizing them (removing all punctuation, stop-words, and lower-casing), then testing for equality. Once we found a pair of equivalent sentences, two cases arise:
\begin{itemize}
    \item The corresponding boxes are highly overlapping (IoU $>$ 0.7). In this case, we consider both annotations to be redundant, and we keep only one of them.
    \item The boxes are non-overlapping. The most likely explanation is that the sentence is under-specified and actually corresponds to several distinct objects in the image. In this case, we merge the two data-points together, and the resulting annotations comports two boxes for this sentence.
\end{itemize}
We iterate recursively this process until no equivalent sentences remain.

In some cases, the VG annotations provide information about the object referred to in the sentence. For example, if the region is tagged ``the cat on the white table", in some cases the individual boxes for the cat and the table are available. In this case, we discard the region box and use the individual boxes instead.
We note that despite our merging strategy, it may remain some region description that do not canonically refer to a unique object in the image, but for which we don’t have ground-truth annotations for the other objects that also match the said region description. Despite the noise it introduces in the training process, we don’t pursue extra efforts to try and fix these situations.
In addition, we also use questions from the GQA dataset \cite{hudson_gqa_2019}, where bounding box annotations are provided for key phrases in the questions.

\begin{figure*}[t!]
    \centering

    \begin{subfigure}[t]{0.45\textwidth}
        \centering
        \includegraphics[height=1.3in]{parts/images/pink.png}
        \label{ele1}
        \caption{Text prompt: ``A pink elephant."}
    \end{subfigure}
    \begin{subfigure}[t]{0.45\textwidth}
        \centering
        \includegraphics[height=1.3in]{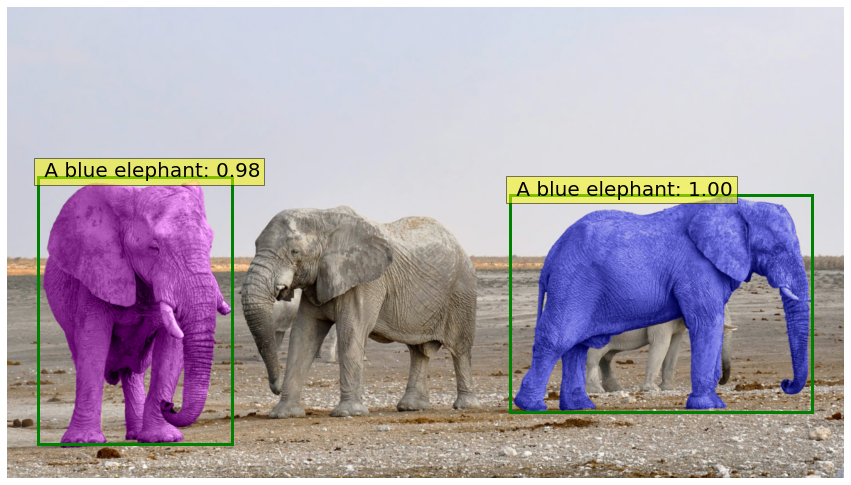}
        \caption{Text prompt: ``A blue elephant."}
    \end{subfigure}
    \\
     \begin{subfigure}[t]{0.45\textwidth}
         \centering
        \includegraphics[height=1.3in]{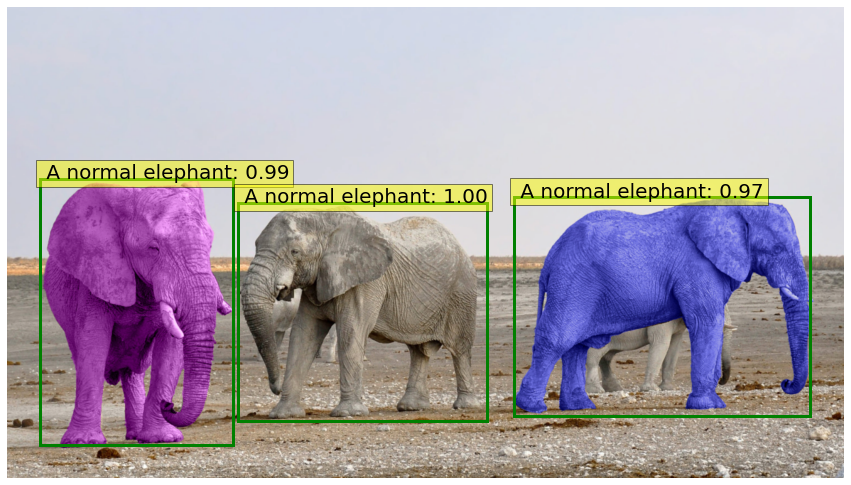}
         \caption{Text prompt: ``A normal elephant."}
     \end{subfigure}
    \begin{subfigure}[t]{0.45\textwidth}
        \centering
        \includegraphics[height=1.3in]{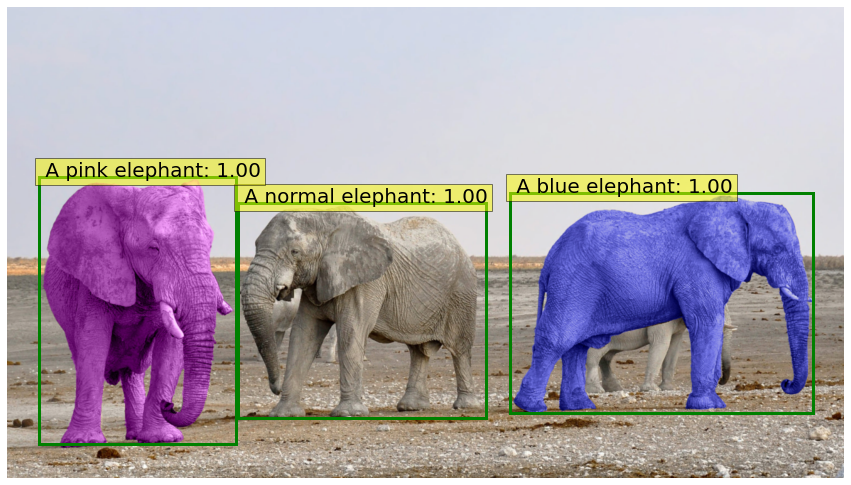}
        \caption{Text prompt: ``A pink elephant. A blue elephant. A normal elephant"}
    \end{subfigure}%
    \caption{Qualitative results on unseen attributes combinations. While the model correctly singles out the pink elephant (a), it incorrectly includes the pink elephant when prompted about the blue one (b). In (c), we show that the model does not understand what a ``normal elephant`` looks like. However, in (d), when prompted about all three elephants at once, it is able to assign the correct label to each of them, by process of elimination.}
\end{figure*}

\section{Evaluating grounded detection}
\label{sec:eval_GD}

The main evaluation metric proposed to evaluate grounded detection in datasets like Flickr30K entities\cite{plummer2015flickr30k} is Recall@$k$, that is measuring whether a given model is able to rank the ``correct'' box amongst the top $k$ it produces. The correctness of a box is decided by computing the Intersection-over-Union (IoU) between the proposed box and the ground-truth box, and deemed correct if the IoU is above a predetermined threshold, generally 0.5.
While this kind of evaluation is well-suited for tasks where there is a clear one-to-one mapping between phrases and boxes, for example in Referring Expression Comprehension tasks, we argue that in general grounded detection tasks, they fall short of properly evaluating the performance of the models. Specifically, they run into the following issues:
\begin{enumerate}
    \item \textbf{Multiple boxes for a given phrase}: Since the recall@k metric implies a single box per phrase, it is not clear how to extend it to situations where a given phrase refers to several distinct objects in the image \nico{TODO image}. In the absence of clear guidelines, various authors have adopted divergent approaches to deal with that. Specifically, some \cite{li2019visualbert,kim2018bilinear} consider the predicted box to be correct if it has an IoU$>0.5$ with \emph{any} of the ground-truth boxes. We refer to this protocol as the \textsc{Any}-Protocol. Others \cite{plummer2020revisiting,yang2019fast} first merge all the ground-truth boxes associated to the phrase by considering the smallest enclosing box. Then they proceed to compute the IoU as usual, using this merged box as the ground-truth one. We refer to this protocol as the \textsc{Merged-Boxes}-Protocol.
    
    Arguably, both methods have drawbacks: the first one keeps the atomicity of each instance but fails to evaluate whether the model found all the referred instances. The second one looses the fine-grained instance in favor of a box that may be unreasonably bloated if the instances are spread apart. %
    
    \item \textbf{Multiple phrases for a given box}. In some cases, the same object is referred to multiple times in the sentence. Sometimes, the corresponding phrases are exact duplicates of each other, or close synonyms (eg ``a guy'' and ``a man''), but sometimes the co-references are more subtle. One such example include referring alternatively to a group (eg ``a couple'') or to a sub-constituent (eg ``the woman'').
    
    Under the current evaluation protocol, each phrase is evaluated independently, even if they refer to the same object. As such, it does not test the model's understanding of co-references, which is arguably an important aspect of learning grounded representations.
    
\end{enumerate}

Recognising the discrepancy in the evaluation procedures in the literature and the difficulties it creates in comparing various approaches, we also evaluate MDETR under the merged-box protocol, as described in Sect~\ref{sec:merged_boxes}.

\subsection{Evaluation under the \textsc{Merged-Boxes}-Protocol}
\label{sec:merged_boxes}
In MDETR, the dataset creation process operates under the assumption that a phrase is associated to all the individual boxes that correspond to it. As a result, MDETR does not naturally predict the merged-boxes that would be required by the \textsc{Merged-Boxes}-Protocol. For that reason, it is necessary to fine-tune the model on a version of Flickr30k entities where the boxes have been merged appropriately. We note that the \textsc{Any}-Protocol did not require such fine-tuning.

\section{Error Analysis}

\begin{figure}[t] \centering
\begin{tikzpicture}
\pie[color={myorange!10, myorange!20, myorange!30, myorange!40, myblue!30, myblue!40, myblue!50, mygreen!40}]{
12/Ambiguous GT location, 
18/Inconsistent annotations (when several objects are referred), 
13/Imprecise GT box, 
15/Wrong GT object,
7/Wrong instance,
14/Wrong object,
2/OCR,
19/Imprecise box prediction
}
\end{tikzpicture}
\caption{Break down of the first 100 errors from the EfficientNetB5 model on the Flickr30k validation set. Yellow shade corresponds to mistakes in the ground truth annotations. Blue shade corresponds to errors made by the model in grounding. Green shade corresponds to errors made by the model in accurate localization. See text for more details.}
\label{fig:errors}
\end{figure}
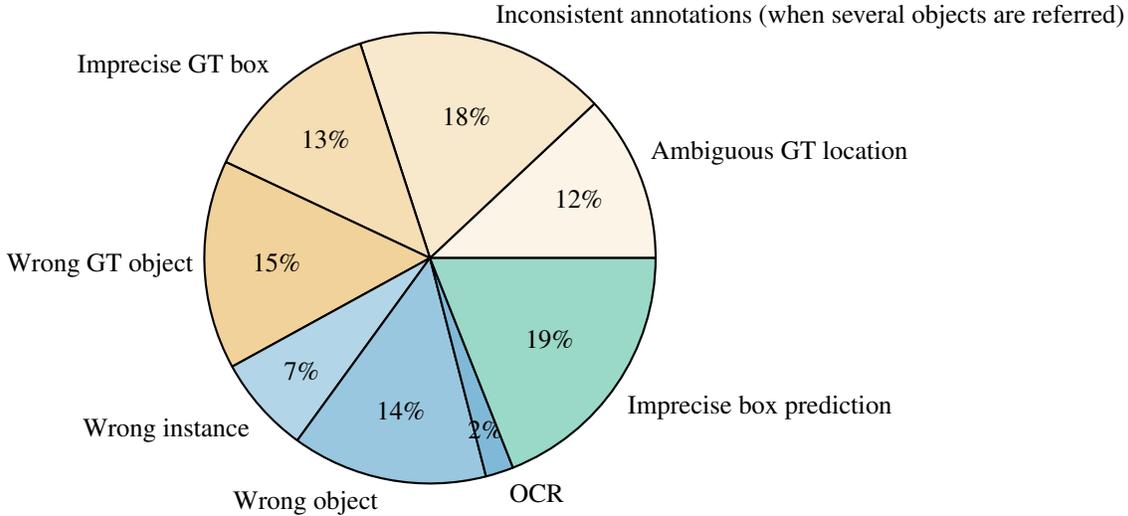

To better understand the failures of the model on the grounding task, we provide a small-scale error analysis. We evaluate our best model, the EfficientNetB5 variant, on the validation set of Flickr30k. We manually inspect the first 100 errors made by the model. The results are summarized in Fig \ref{fig:errors} and we detail the types of errors we uncovered in the following:
\begin{itemize}
    \item Issues with the ground truth annotations
    \begin{description}
\item [Ambiguous GT location] This corresponds to phrases that don't have a canonical localization. They mostly correspond to scene elements, such as ``beach" or ``tree".
\item [Inconsistent annotations (when several objects are referred)] In the \textsc{Any}-Protocol, if a phrase corresponds to several objects, then in principle every single instance should be individually annotated. However, we find some inconsistencies in the annotations, for example some instances that are missed, or some distinct instances that are annotated using the same box.
\item [Imprecise GT box] This corresponds to cases where the provided box is not adequate. It is either too big (not tight), or too small, cutting out a part of the object.
\item[Wrong GT] In those cases, the annotated box(es) don't correspond to the correct referred object at all.
\end{description}
\item Grounding mistakes
    \begin{description}
\item [Wrong instance] The model picks an instance of the correct type (including adjective modifiers, if any), however it is not the correct instance when taking into account context from the rest of the sentence. 
\item[Wrong object] The model picks the wrong object, and it is not of the correct type. This usually occurs on long-tail concepts that the model doesn't seem to know about.
\item[OCR] The phrase refers to written text, thus requiring OCR abilities from the model, which it doesn't have.
\end{description}
\item Detection issues
    \begin{description}
\item [Imprecise box prediction] The model is clearly selecting the right object, however the predicted box isn't quite precise enough. This happens on small objects as well as elongated objects, where a relatively small L1 error can cause a low IoU with the ground truth.

\end{description}
\end{itemize}

Overall, we find that on the analyzed subset, more than half the errors stem from issues in the ground-truth annotations. Extrapolating to the rest of the dataset, this would imply a label noise of around 10\%, which might be detrimental to make significant further progress on the task.

\section{Experiments on VQA2}
\label{sec:vqa2}
In our visual question answering experiments on the GQA dataset, we always had access to bounding box information for the questions. Only during fine-tuning on the balanced set for 10 epochs, we do not supervise the detection losses. On the other hand, for datasets such as VQA2~\cite{balanced_vqa_v2}, we do not have access to any box annotations and the supervision comes solely from the question answering loss. We fine-tune two of our models --- pre-trained on the joint dataset, as well as the model fine-tuned on the all-split of GQA --- on the VQA v2 dataset for 25 epochs. The results are reported in Table~\ref{tab:vqa2}.  
While these results are not state-of-the-art on the VQA2 benchmark, they are still quite reasonable with respect to current literature. This show that our method can be extended to tasks where we do not have the dense supervision (in the form of bounding boxes and their alignment to the text) that we otherwise expect in tasks reported in this paper.  

\begin{table}[h]
\begin{center}
\small
\begin{tabular}{lcccccccc}
 \toprule
 Pre-training & \multicolumn{4}{c}{Test-Dev} & \multicolumn{4}{c}{Test-Std}\\
              & Overall & Yes-no & Other & Number & Overall & Yes-no & Other & Number \\
 \midrule
Modulated detection on combined dataset (40 epochs) & 70.49 & 86.74 & 55.17 & 60.04 & - & - & - & -\\
+ GQA-\textit{all} (5 epochs) & 70.64 & 86.74 & 55.26 & 60.33 & 70.63 & 86.79 & 55.12 & 60.12  \\
 \bottomrule
\end{tabular}
\caption{VQA v2 results
\label{tab:vqa2}}
\end{center}
\end{table}

\end{document}